\documentclass{fairmeta}
\usepackage{amsmath}
\DeclareMathOperator*{\argmax}{arg\,max}
\usepackage{enumerate} 
\usepackage{bm}
\usepackage{algorithm}
\usepackage{floatflt}
\usepackage{algpseudocode}
\usepackage{amsfonts}
\usepackage{amsthm}
\usepackage{newtxtt}
\usepackage{algorithm}
\usepackage{colortbl} 
\usepackage{cleveref}
\usepackage{diagbox} 
\usepackage[utf8]{inputenc}
\usepackage{textgreek}
\usepackage{colortbl}
\usepackage{nicematrix}
\usepackage{makecell}
\usepackage{float}
\usepackage[frozencache,cachedir=.]{minted}
\usepackage{caption}
\usepackage{subcaption}
\usepackage[table]{xcolor}
\usepackage{tcolorbox}
\usepackage{amssymb}
\usepackage{xspace}
\usepackage{wrapfig}
\usepackage{adjustbox}
\usepackage{tabularx}
\usepackage{booktabs}
\usepackage{mathtools}
\usepackage{wrapfig}
\usepackage{amssymb}
\usepackage{graphicx}
\usepackage{pifont}
\usepackage{hyperref}
\usepackage{url}
\usepackage{multirow}
\usepackage{array}

\newcolumntype{C}{>{\centering\arraybackslash}X}

\definecolor{upColor}{RGB}{17,138,21}
\definecolor{downColor}{RGB}{174,36,67}

\newtheorem{theorem}{Theorem}[]

\newtheorem{remark1}[theorem]{Remark}

\title{ReferFlow: Beyond Final-Target Localization for Complex Referring Expressions}
\author[1,2,\textsuperscript{$\dagger$}]{Tianyi Gao}
\author[2,\textsuperscript{$\dagger$}]{Han Fang}
\author[2]{Xin Wei}
\author[3]{Yaoyao Zhong}
\author[2,4]{Zhaoyan Pan}
\author[2,5]{Weijie Zhu}
\author[2]{Hongbo Sun}
\author[2]{Zhiyu Lin}
\author[2]{Kaicheng Yang}
\author[6]{Jinglin Xu}
\author[2]{Ye Yuan}
\author[7]{Hao Li}
\author[2,*]{Hao Sun}
\author[1,*]{Jingmin Xin}
\affiliation[1]{State Key Laboratory of Human-Machine Hybrid Augmented Intelligence, Institute of Artificial Intelligence and Robotics, Xi'an Jiaotong University}
\affiliation[2]{Xingchen AGI Lab, China Telecom Artificial Intelligence Technology (Beijing) Co., Ltd.}
\affiliation[3]{Beijing Normal University}
\affiliation[4]{Zhejiang University}
\affiliation[5]{University of the Chinese Academy of Sciences}
\affiliation[6]{University of Science and Technology Beijing}
\affiliation[7]{Shanghai Jiao Tong University}
\metadata[Links]{ \href{https://v-insight-ai.github.io/ReferFlow/}{Project Page} $|$ \href{https://github.com/V-Insight-AI/ReferFlow}{GitHub} }
\date{Oct 14, 2026}
\begin{document}

\abstract{
  Multimodal Large Language Models have made significant progress in referring expression comprehension (REC), yet most REC benchmarks still focus on final-target localization. Complex queries often involve intermediate visual references with different dependencies on prior grounding, but final-only evaluation leaves this process unexamined. To address this gap, we formulate trajectory-level REC, in which models receive an ordered sequence of grounding steps constructed from a single complex query and localize the target regions associated with each step. 
  We instantiate this setting in ReferFlow-Bench, which characterizes structured region transitions through four trajectory types, namely Refinement, Shift, Reset, and Mixture.
  Experiments show that current models struggle to consistently ground intermediate visual references across successive steps.
  We further propose ReferFlow, which decouples interleaved text–latent trajectory generation from coordinate decoding. 
  Specifically, Stateful Region Latent Reasoning represents each target region as an object-indexed state composed of continuous latents, with explicit visual and spatial supervision to preserve region information for subsequent grounding.
  Trajectory-Aware Parallel Decoding then recovers all coordinate blocks in parallel through four geometry-constrained refinement rounds, with each block conditioned on its corresponding trajectory prefix, thereby reducing within-box sequential dependencies and eliminating cross-box autoregression.
  We further apply Grounding-Aware GRPO to promote consistent grounding across steps and learn adaptive trajectory generation from complex REC queries.
   ReferFlow achieves leading performance on general REC benchmarks and ReferFlow-Bench.
}

\maketitle

\begingroup
\renewcommand{\thefootnote}{}
\footnotetext{%
  \rmfamily\bfseries
  \textsuperscript{$\dagger$}Equal Contributions
  \quad
  \textsuperscript{*}Corresponding Authors
}
\addtocounter{footnote}{-1}
\endgroup

\section{Introduction}

\begin{figure}
  \includegraphics[width=1\linewidth]{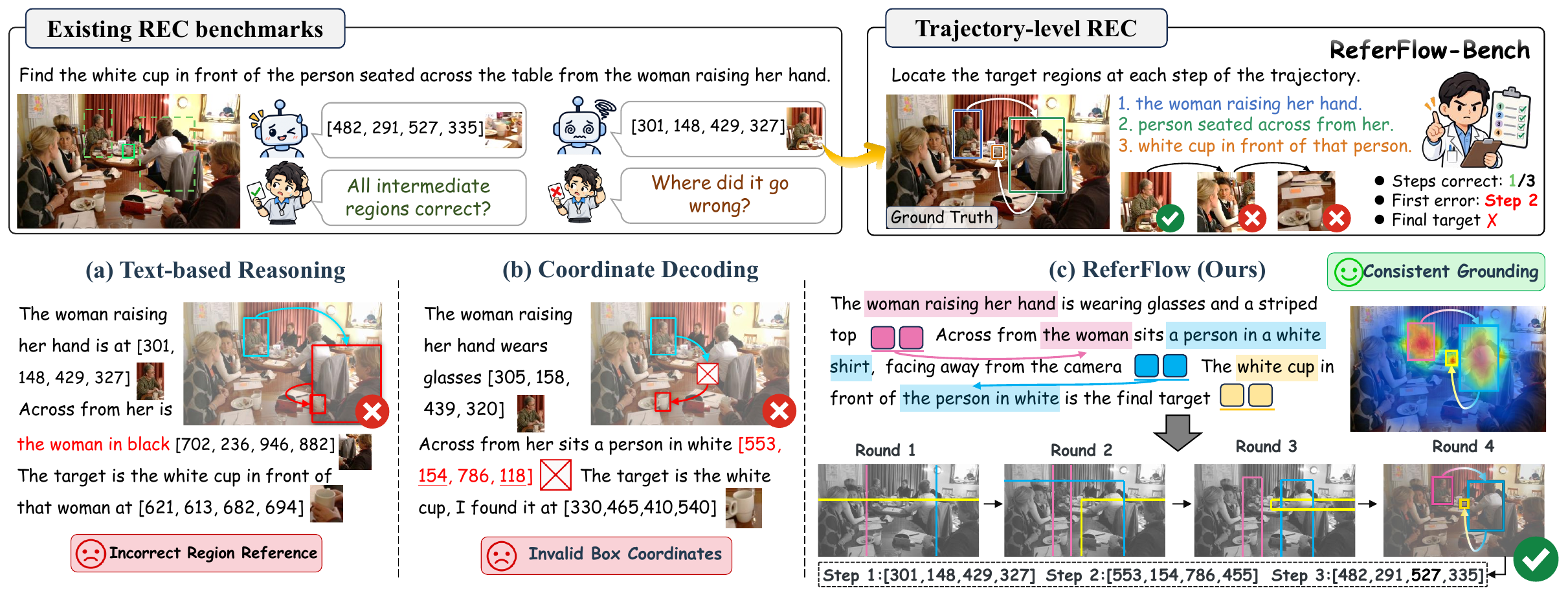}
  \caption{
ReferFlow-Bench evaluates grounding trajectories, exposing intermediate localization failures hidden by final-only REC evaluation.
ReferFlow combines object-indexed region states with parallel geometry-constrained coordinate decoding to support consistent grounding across steps.
  }
  \label{fig:motivation}
\end{figure}

Multimodal Large Language Models have made significant progress in referring expression comprehension (REC), with many early REC benchmarks becoming saturated~\citep{kazemzadeh2014referitgame}. 
Recent work has explored more complex visual grounding with more challenging queries, including detailed descriptions, hard distractors, fine-grained visual understanding, and compositional reasoning~\citep{chen2025revisiting,dong2026ref,li2026groundingme,gao2026mixture}. 
As shown in Fig.~\ref{fig:motivation}, such queries often specify a target through other objects and their relationships, bringing multiple visual references into the grounding process. Across successive grounding steps, some references build on previously localized regions, while others introduce independent targets. However, most REC benchmarks still focus on final-target localization, leaving the regions associated with these intermediate references unexamined. 
While recent multi-turn grounding methods extend evaluation beyond a single prediction, they mainly study how region references are propagated as new queries are introduced across dialogue turns~\citep{sun2026regionreasoner}. In contrast, localization under different cross-step region dependencies within a single query has not been systematically evaluated.

To address this gap, we formulate trajectory-level REC: given an image and an ordered sequence of grounding steps constructed from a single complex query, models localize the target regions associated with each step.
We instantiate this setting in ReferFlow-Bench, which differs from existing benchmarks in three respects:
(1) \textbf{Intermediate grounding trajectory.} ReferFlow-Bench organizes intermediate visual references into an ordered grounding trajectory, making their localization observable.
(2) \textbf{Structured region transitions.} Four trajectory types characterize how target regions and their dependencies change across steps: Refinement, Shift, Reset, and Mixture. These trajectories require models to refine existing targets, use prior regions to locate related targets, and localize independent targets while avoiding interference from earlier grounding.
(3) \textbf{Trajectory-level evaluation.} Each grounding step is paired with target-region annotations, enabling evaluation of both individual steps and the entire trajectory.
Evaluating more than 20 MLLMs reveals substantial difficulty in consistently grounding intermediate visual references throughout the trajectory.

These results highlight two limitations of existing MLLM-based grounding methods (Fig.~\ref{fig:motivation}).
First, tracking intermediate regions and reasoning over their visual content and spatial information across grounding steps remains challenging. Chain-of-Thought reasons over complex conditions in language~\citep{jiang2026rex,DBLP:conf/nips/FanHYZKZGW25}, while Thinking-with-Images accesses visual evidence through predicted coordinates~\citep{zheng2026deepeyes}. These external references make it difficult to capture fine-grained visual and spatial cues for subsequent grounding.
Second, most MLLMs autoregressively generate bounding boxes as coordinate tokens~\citep{bai2025qwen3}. Parallel Box Decoding jointly predicts coordinates within each box but retains autoregressive dependencies across boxes~\citep{wang2026locateanything}. Thus, coordinate-decoding depth grows with box count, while small coordinate-token errors can substantially shift predicted regions and mislead subsequent steps.

To address these issues, we propose ReferFlow, which decouples interleaved text-latent trajectory generation from coordinate decoding. 
 Stateful Region Latent Reasoning (SRLR) represents each target with an object-indexed region state composed of continuous latents.
An auxiliary decoder supervises each state with target-region features and boxes to retain fine-grained visual and spatial information, allowing the model to determine the relevance of prior regions to the current step.
After trajectory generation, Trajectory-Aware Parallel Decoding (TAPD) decodes all target boxes in parallel over four geometry-constrained refinement rounds, making coordinate-decoding depth independent of the number of boxes. A trajectory-aware mask isolates coordinate blocks while conditioning each on its corresponding trajectory prefix. Within each block, TAPD treats multi-digit coordinates as atomic units and recovers the highest-confidence geometrically feasible coordinate at each round.
We further apply Grounding-Aware GRPO to estimate intermediate-step values from the final localization outcomes of prefix rollouts.
This promotes consistent grounding across steps and adaptive trajectory generation  through task-relevant regions.
Our main contributions are:

\begin{itemize}
\item We formulate trajectory-level REC and construct ReferFlow-Bench for structured evaluation of grounding under different dependencies on prior regions. Ordered grounding steps and target-region annotations enable step-level and trajectory-level evaluation.
\item We propose ReferFlow, which maintains object-indexed region states under visual and spatial supervision and decodes all corresponding boxes in parallel through geometry-constrained refinement. We further apply Grounding-Aware GRPO to learn adaptive trajectories directly from complex queries.
\item We evaluate more than 20 MLLMs, revealing difficulties in consistently grounding intermediate visual references. Extensive experiments show that ReferFlow achieves leading performance on general REC benchmarks and ReferFlow-Bench.
\end{itemize}

\section{Related Work}

\subsection{Referring Expression Comprehension Benchmarks}

RefCOCO, RefCOCO+~\citep{kazemzadeh2014referitgame}, and RefCOCOg~\citep{yu2016modeling} have served as foundational REC benchmarks.
Recent benchmarks introduce more challenging queries and visual conditions.
Ref-L4~\citep{chen2025revisiting} uses long referring expressions, while Ref-Adv~\citep{dong2026ref} focuses on visually similar distractors.
GroundingME~\citep{li2026groundingme}, PR-Bench~\citep{gao2026mixture}, and HumanRef~\citep{jiang2025referring} examine fine-grained visual understanding and reasoning.
Despite this increased complexity, most REC benchmarks still focus on final-target localization.
RegionDial-Bench~\citep{sun2026regionreasoner} and VRT-Bench~\citep{yuan2025visual} extend evaluation beyond a single prediction through multi-turn reference grounding and textual reasoning with object grounding, respectively.
However, they do not separately evaluate distinct types of cross-step region dependencies.
ReferFlow-Bench instead provides ordered grounding steps with structured region transitions from a single complex query for step-level and trajectory-level evaluation.

\subsection{MLLMs in Grounding Tasks}
MLLM-based methods improve visual grounding through advances in reasoning and coordinate decoding.
Rex-Thinker~\citep{jiang2026rex} uses chain-of-thought (CoT) to analyze candidate regions, while GRIT~\citep{DBLP:conf/nips/FanHYZKZGW25} interleaves textual reasoning with region coordinates.
These references may not fully capture fine-grained visual and spatial information.
DeepEyes~\citep{zheng2026deepeyes} accesses visual evidence through region crops but requires additional visual processing.
For coordinate decoding, Rex-Omni~\citep{jiang2026detect} generates quantized coordinates via next-token prediction (NTP), while LocateAnything~\citep{wang2026locateanything} uses multi-token prediction (MTP) for parallel within-box decoding, with NTP fallback to reduce spatial ambiguity.
However, both still rely on autoregressive decoding across boxes.
ReferFlow instead learns object-indexed region states with visual and spatial supervision for cross-step grounding. It decouples trajectory generation from coordinate decoding, recovering all boxes in parallel through geometry-constrained refinement.

\section{ReferFlow-Bench}

\textbf{Problem Formulation.}
We formulate trajectory-level REC to evaluate whether models can consistently ground visual references under different cross-step region dependencies. Given an image $I$ and an ordered sequence of $T$ grounding steps $\mathcal{S}=(s_1,s_2,\ldots,s_T)$ constructed from a single complex query, the model predicts the target boxes associated with each step:
\begin{equation}
(I, \mathcal{S}) \longrightarrow (\mathcal{B}_1, \mathcal{B}_2, \ldots, \mathcal{B}_T),
\label{eq:problem_formulation}
\end{equation}
where $\mathcal{B}_t$ denotes the set of target boxes for step $s_t$. As the trajectory progresses, models must refine existing targets, use prior regions to locate related targets, or localize independent targets. Based on this formulation, we construct ReferFlow-Bench  covering diverse region transitions and providing target-region annotations for every step. The construction pipeline is provided in the Appendix~\ref{annotation_pipeline}.

\begin{figure}
\includegraphics[width=\linewidth]{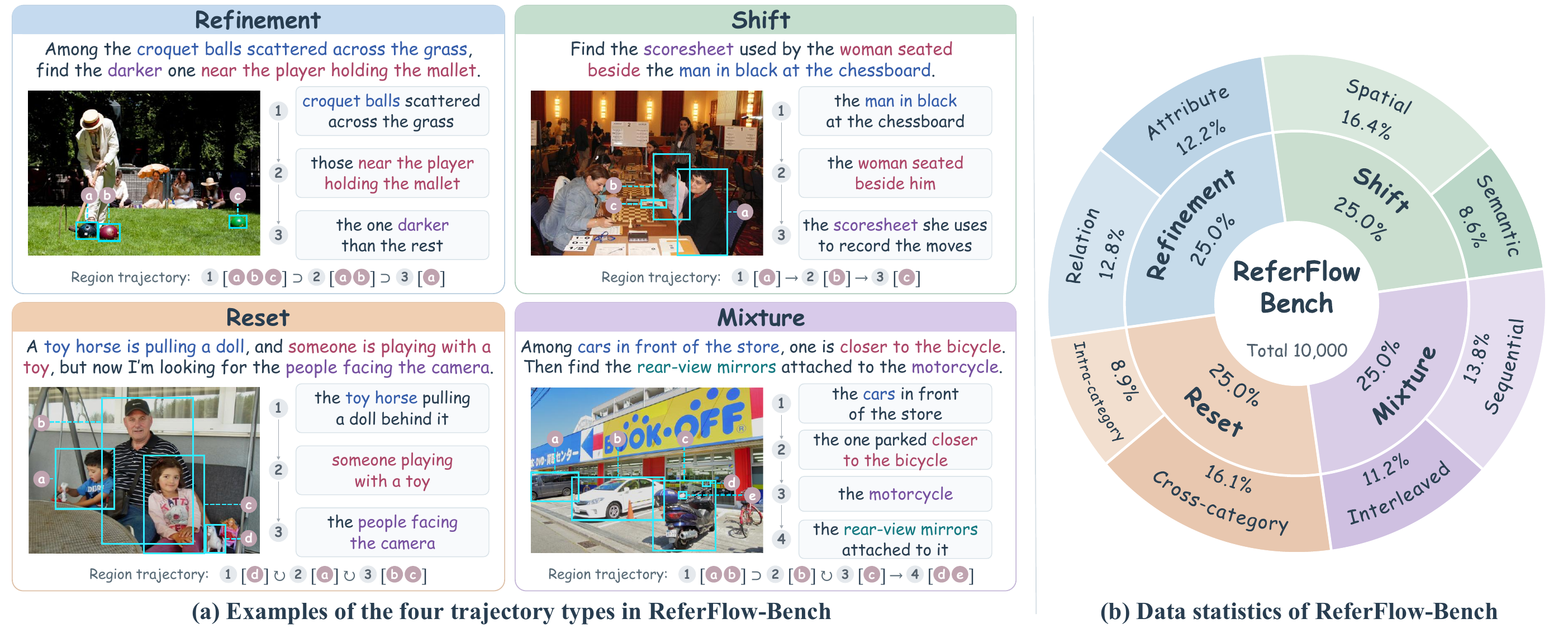}
\caption{\textbf{Overview of ReferFlow-Bench.} (a) Four examples illustrate the trajectory types with different cross-step region dependencies. (b) Distribution of the 10,000 constructed trajectories.}
\label{fig:bmk}
\end{figure}

\textbf{Types of Trajectories.}
As shown in Fig.~\ref{fig:bmk}, ReferFlow-Bench defines four trajectory types according to how target regions and their dependencies change across steps: Refinement, Shift, Reset, and Mixture. 
Each trajectory type is further divided into two subtypes, as detailed below:
\begin{enumerate}
\item \textbf{\textit{Refinement}} requires models to narrow an existing target set by accumulating constraints. Attribute refinement adds constraints on color and other visual properties, whereas relational refinement adds spatial, interaction, or contextual constraints involving other objects.

\item \textbf{\textit{Shift}} requires models to use previously localized regions to identify related targets. Spatial shift is based on relative position or containment, whereas semantic shift is defined by interaction, ownership, part--whole, or other semantic relations.

\item \textbf{\textit{Reset}} requires models to localize independently specified targets while avoiding interference from prior grounding. Cross-category reset switches to a different category, whereas intra-category reset switches to another instance of the same category.

\item \textbf{\textit{Mixture}} combines different types of region transitions within a trajectory. In sequential mixture, each subsequent step builds on the target regions referenced in the immediately preceding step. Interleaved mixture includes at least one subsequent step that depends on regions from a non-adjacent earlier step or requires independent localization.
\end{enumerate}

\textbf{Evaluation Settings.}
We adopt Precision\text{@}($F_1=1$, IoU$\geq0.5$) as the grounding criterion. A step is correct only when the predicted and ground-truth box sets admit a one-to-one matching covering both sets, with IoU$\geq0.5$ for every matched pair. ReferFlow-Bench reports three metrics. Query-level Accuracy (Q-Acc) averages the proportion of correctly grounded steps across trajectories. Trajectory-level Accuracy (T-Acc) measures the proportion of trajectories in which every step is correct. Final-step Accuracy (F-Acc) measures accuracy at the final step.

\section{Method}
\label{sec:method}

\subsection{Stateful Region Latent Reasoning}

\begin{figure}
\includegraphics[width=\linewidth]{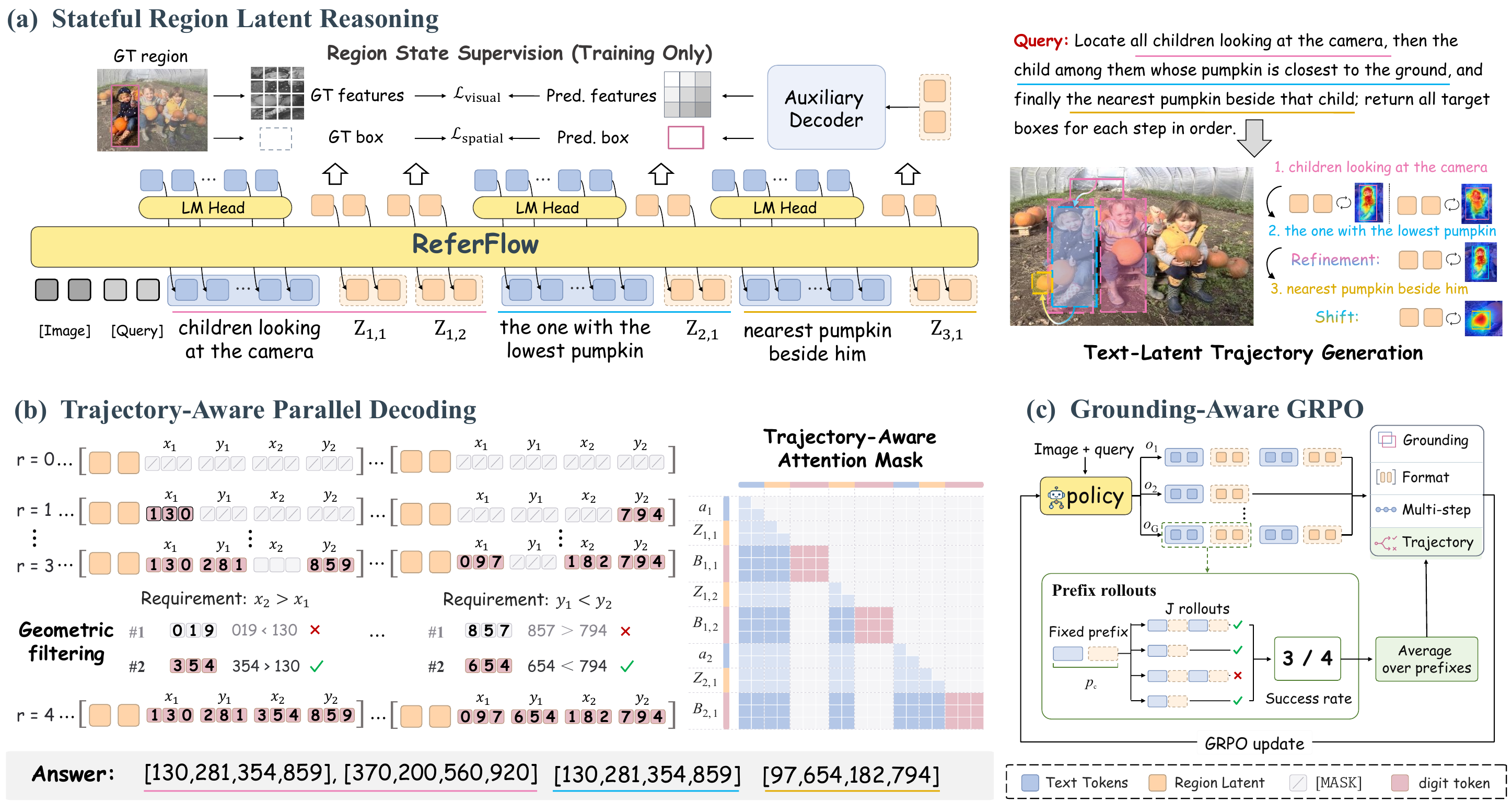}
\caption{Overview of ReferFlow. ReferFlow builds an interleaved text-latent grounding trajectory with Stateful Region Latent Reasoning, then uses Trajectory-Aware Parallel Decoding to decode all target boxes in parallel from their corresponding trajectory prefixes through geometry-constrained refinement. Grounding-Aware GRPO learns adaptive trajectories for complex REC.}

\label{fig:method}
\end{figure}

As illustrated in Fig.~\ref{fig:method}, SRLR represents each target with an object-indexed region state composed of continuous latents. Each state is explicitly supervised to preserve the target's visual content and spatial information for subsequent grounding.
Given an image $I$ and grounding steps $\mathcal{S}$, SRLR interleaves textual responses and region states into a structured text-latent trajectory $\mathcal{H}=[a_1,\mathcal{Z}_1,\ldots,a_T,\mathcal{Z}_T]$, where $a_t$ is the textual response at step $t$ and $\mathcal{Z}_t$ comprises the corresponding $N_t$ region states:
\begin{equation}
\mathcal{Z}_t = \left( \mathbf{Z}_{t,1}, \ldots, \mathbf{Z}_{t,N_t} \right), \qquad
\mathbf{Z}_{t,n} = [\mathbf{z}_{t,n,1},\ldots,\mathbf{z}_{t,n,M}] \in \mathbb{R}^{M\times d},
\label{eq:region_state_def}
\end{equation}
where $\mathbf{Z}_{t,n}$ is the region state of the $n$-th target and consists of $M$ region latents with hidden dimension $d$.
Each region state is delimited by \texttt{<|latent\_start|>} and \texttt{<|latent\_end|>}, and its $m$-th latent is generated in the hidden space as follows:
\begin{equation}
\mathbf{z}_{t,n,m} = f_{\theta}^{\mathrm{hidden}} \left( I, \mathcal{S}, \mathcal{H}_{<t}, a_t, \mathcal{Z}_{t,<n}, \mathbf{Z}_{t,n,<m} \right),
\label{eq:z_generation}
\end{equation}
where $\mathcal{H}_{<t}$, $\mathcal{Z}_{t,<n}$, and $\mathbf{Z}_{t,n,<m}$ 
denote the preceding text-latent trajectory, region states at step $t$,
and latents within the current region state, respectively.
Rather than externalizing intermediate region representations as discrete coordinate tokens, SRLR feeds the generated latents directly back as continuous input embeddings. 
These region states remain in $\mathcal{H}$ as context for subsequent grounding.
To associate each state with the content and location of its target region, we jointly supervise it using target-region visual features and bounding boxes. An auxiliary decoder reconstructs the visual features $\hat{\mathbf{F}}_{t,n}$ through cross-attention and predicts the bounding box $\hat{\mathbf{b}}_{t,n}$ through an MLP:
\begin{equation}
\hat{\mathbf{F}}_{t,n} = \operatorname{Attn} \left( \mathbf{P}, \mathbf{Z}_{t,n}, \mathbf{Z}_{t,n} \right), \qquad
\hat{\mathbf{b}}_{t,n} = \operatorname{Sigmoid} \left( \operatorname{MLP}_{b} \left( [\mathbf{z}_{t,n,1};\ldots;\mathbf{z}_{t,n,M}] \right) \right),
\label{eq:aux_decoder}
\end{equation}
where $\mathbf{P}$ denotes positional query embeddings constructed using two-dimensional sine-cosine encoding. Feature reconstruction encourages each state to retain target-specific visual content, while box prediction constrains its spatial correspondence to the target region. These jointly supervised states provide context for subsequent grounding and parallel coordinate decoding in TAPD. The auxiliary decoder is used only during training and removed at inference time.

\subsection{Trajectory-Aware Parallel Decoding}

Given the text-latent trajectory generated by SRLR, TAPD instantiates a coordinate block for each region state and decodes all blocks in parallel. Each block uses its corresponding trajectory prefix as context without depending on the decoded coordinates of other boxes.
We quantize each box coordinate to a value in $\{0,\ldots,999\}$ and encode it using three decimal digit tokens, resulting in a fixed-length coordinate block $\mathcal{B}_{t,n}=(b_{t,n,1},\ldots,b_{t,n,L_b})$ with $L_b=12$. Each coordinate block $\mathcal{B}_{t,n}$ is inserted after its corresponding region state $\mathbf{Z}_{t,n}$ to obtain the expanded trajectory:
\begin{equation}
\widetilde{\mathcal{H}} =
\left[
a_1,
(\mathbf{Z}_{1,n},\mathcal{B}_{1,n})_{n=1}^{N_1},
\ldots,
a_T,
(\mathbf{Z}_{T,n},\mathcal{B}_{T,n})_{n=1}^{N_T}
\right].
\label{eq:expanded_trajectory}
\vspace{-0.2cm}
\end{equation}

Coordinate blocks are initialized with \texttt{[MASK]} tokens and decoded over $R$ rounds. 
The trajectory-aware mask preserves causal attention within the text-latent trajectory and excludes coordinate blocks from its context. Each block $\mathcal{B}_{t,n}$ attends to the text-latent prefix ending at its corresponding region state $\mathbf{Z}_{t,n}$, while remaining isolated from all other coordinate blocks.
Within each block, bidirectional attention allows all masked digit tokens to be predicted in parallel. At round $r$, $\mathcal{M}_{t,n}^{(r)}$ denotes the masked positions in $\mathcal{B}_{t,n}$. The masked-token distribution factorizes as:
\begin{equation}
p_\theta
\left(
\mathcal{B}_{\mathrm{mask}}
\mid
\widetilde{\mathcal{H}}^{(r)}
\right)
=
\prod_{t=1}^{T}
\prod_{n=1}^{N_t}
\prod_{i\in\mathcal{M}_{t,n}^{(r)}}
p_\theta
\left(
b_{t,n,i}
\mid
\mathcal{H}_{\leq(t,n)},
\mathcal{B}_{t,n}^{(r)}
\right),
\label{eq:parallel_factorization}
\vspace{-0.1cm}
\end{equation}
where $\widetilde{\mathcal{H}}^{(r)}$ and $\mathcal{B}_{t,n}^{(r)}$ denote the expanded trajectory and coordinate block at round $r$, $\mathcal{H}_{\leq(t,n)}$ denotes the text-latent prefix ending at $\mathbf{Z}_{t,n}$, and $\mathcal{B}_{\mathrm{mask}}$ denotes all masked coordinate tokens.

TAPD treats each coordinate as an atomic unit by jointly scoring its three digit tokens, avoiding a dedicated token for every quantized value.
Within each block, we omit the indices $(t,n)$ and denote the unrecovered coordinates by $\mathcal{C}_r$. For each $c\in\mathcal{C}_r$, previously recovered coordinates determine a feasible interval $\mathcal{I}_r(c)$ consistent with box constraints $x_{\min}<x_{\max}$ and $y_{\min}<y_{\max}$. The feasible candidate set is $\mathcal{V}_r(c)=\mathcal{I}_r(c)\cap\{0,\ldots,999\}$.
Let $p_{r,c,j}$ denote the predicted digit distribution at the $j$-th position of coordinate $c$. 
 Each $v\in\mathcal{V}_r(c)$ is represented by the zero-padded digit sequence $(d_1(v),d_2(v),d_3(v))$. We score all feasible candidates and select one coordinate as follows:
\begin{equation}
S_r(c,v)=\sum_{j=1}^{3}\log p_{r,c,j}(d_j(v)),\quad
\hat{v}_r(c)=\argmax_{v\in\mathcal{V}_r(c)}S_r(c,v),\quad
c_r^*=\argmax_{c\in\mathcal{C}_r}S_r(c,\hat{v}_r(c)).
\label{eq:coordinate_selection}
\end{equation}
The selected value $\hat{v}_r(c_r^*)$ is recovered as a three-token unit, while the remaining coordinates stay masked with updated feasible intervals.
Since each block recovers one coordinate per round, all boxes are decoded synchronously in $R=4$ geometry-constrained rounds. TAPD thus eliminates cross-box autoregression and makes coordinate-decoding depth independent of the number of boxes.

\subsection{Grounding-Aware Training Strategy}
We adopt a two-stage training scheme with SFT followed by Grounding-Aware GRPO.
During SFT, SRLR and TAPD are jointly trained on REC ($T=1$) and trajectory-level REC data.
We also pair original complex queries with annotated grounding trajectories to supervise trajectory generation.
\textbf{Supervised Fine-Tuning.}
We jointly optimize text generation, region-state learning, and coordinate decoding. For region-state learning, we combine visual-feature reconstruction and box supervision:
\begin{equation}
\mathcal{L}_{\mathrm{region}} =
\frac{1}{N_{\mathrm{region}}}
\sum\nolimits_{t=1}^{T}
\sum\nolimits_{n=1}^{N_t}
\left(
\lambda_{\mathrm{visual}}\mathcal{L}_{\mathrm{visual}}^{t,n}
+
\lambda_{\mathrm{spatial}}\mathcal{L}_{\mathrm{spatial}}^{t,n}
\right),
\label{eq:loss_region}
\end{equation}
where $N_{\mathrm{region}}=\sum_{t=1}^{T}N_t$. $\mathcal{L}_{\mathrm{visual}}^{t,n}$ combines cosine-similarity and MSE losses between $\hat{\mathbf{F}}_{t,n}$ and the reference features $\mathbf{F}_{t,n}^{*}$ extracted from the ground-truth region by the frozen vision encoder, while $\mathcal{L}_{\mathrm{spatial}}^{t,n}$ combines $L_1$ and GIoU losses for bounding-box supervision.

For TAPD, we randomly mask coordinates as three-token units in each ground-truth block and train the model to recover them from the trajectory prefix and the resulting block $\widetilde{\mathcal{B}}_{t,n}$:
\begin{equation}
\mathcal{L}_{\mathrm{box}} =
- \mathbb{E}_{\mathcal{M}_{t,n}}
\left[
\frac{1}{N_{\mathrm{mask}}}
\sum\nolimits_{t=1}^{T}
\sum\nolimits_{n=1}^{N_t}
\sum\nolimits_{i\in\mathcal{M}_{t,n}}
\log
p_{\theta}
\left(
b_{t,n,i}
\mid
\mathcal{H}_{\leq(t,n)},
\widetilde{\mathcal{B}}_{t,n}
\right)
\right],
\label{eq:loss_box}
\end{equation}
where $\mathcal{M}_{t,n}$ denotes the masked positions, and $N_{\mathrm{mask}}$ denotes the total number of masked positions across the trajectory. Combined with the text modeling loss $\mathcal{L}_{\mathrm{txt}}$, the SFT objective is:
\begin{equation}
\mathcal{L}_{\mathrm{SFT}} =
\lambda_{\mathrm{txt}}\mathcal{L}_{\mathrm{txt}}
+
\lambda_{\mathrm{region}}\mathcal{L}_{\mathrm{region}}
+
\lambda_{\mathrm{box}}\mathcal{L}_{\mathrm{box}}.
\label{eq:loss_sft}
\end{equation}

\textbf{Grounding-Aware GRPO.}
Given an image and either a referring expression or an ordered sequence of grounding steps, we sample $G$ trajectories.
At supervised step $t$ of trajectory $o$, $\hat{\mathcal{Y}}$ and $\mathcal{Y}^{*}$ denote predicted and ground-truth box sets.
IoU-based Hungarian matching $\Pi$ gives the step reward:
\begin{equation}
  R_{\mathrm{gnd}}^{t}(o)=
  \sum\nolimits_{(i,j)\in\Pi}
  \mathrm{IoU}(\hat{\mathbf{b}}_i,\mathbf{b}_j^{*})
  \Big/
  \max(|\hat{\mathcal{Y}}|,|\mathcal{Y}^*|,1).
\label{eq:reward_gnd}
\end{equation}
We set $R_{\mathrm{gnd}}(o)$ to the final-step $R_{\mathrm{gnd}}^{t}(o)$ for REC and its mean over provided steps for trajectory-level REC. The format reward $R_{\mathrm{fmt}}(o)$ is $1$ for interleaved outputs with exactly $M$ latents per state and $0$ otherwise.
For a $K(o)$-step trajectory $o$, we estimate prefix values for the trajectory reward $R_{\mathrm{TR}}$ using $J$ completed rollouts $o_c^{(j)}$ from each prefix $p_c$:
\begin{equation}
p_c=[a_1,\mathcal{Z}_1,\ldots,a_c,\mathcal{Z}_c], \qquad
\widehat{V}(p_c)=
\frac{1}{J}
\sum\nolimits_{j=1}^{J}
S_{\mathrm{final}}\left(o_c^{(j)}\right),
\qquad 1\leq c<K(o),
\label{eq:prefix_value}
\end{equation}
where $S_{\mathrm{final}}(o)=1$ if the final grounding is correct and $0$ otherwise. We define $R_{\mathrm{TR}}$ as:
\begin{equation}
R_{\mathrm{TR}}(o)=
\begin{cases}
\frac{1}{K(o)-1}
\sum\nolimits_{c=1}^{K(o)-1}
\widehat{V}(p_c),
& \text{if } K(o)\geq 2, \\
0,
& \text{otherwise}.
\end{cases}
\label{eq:reward_tr}
\end{equation}

To encourage successful multi-step grounding, we define $R_{\mathrm{multi}}(o)=S_{\mathrm{final}}(o)$ when $K(o)\geq 2$, and $0$ otherwise. The overall reward is:
\begin{equation}
R(o)=
\lambda_{\mathrm{fmt}}R_{\mathrm{fmt}}(o)
+\lambda_{\mathrm{gnd}}R_{\mathrm{gnd}}(o)
+\lambda_{\mathrm{TR}}R_{\mathrm{TR}}(o)
+\lambda_{\mathrm{multi}}R_{\mathrm{multi}}(o).
\label{eq:reward_total}
\end{equation}

\section{Experiments}

\subsection{Implementation Details}\label{sec:setting}
ReferFlow is built upon Qwen3-VL-2B~\citep{bai2025qwen3}. During SFT, we train on 1.6M samples for one epoch with a global batch size of 64 and a learning rate of $1\times10^{-5}$. The loss weights are set to $\lambda_{\mathrm{txt}}=1.0$, $\lambda_{\mathrm{region}}=\lambda_{\mathrm{box}}=0.8$, $\lambda_{\mathrm{visual}}=0.2$, and $\lambda_{\mathrm{spatial}}=0.3$.
During Grounding-Aware GRPO, we train on 13K samples for one epoch with a global prompt batch size of 8 and a learning rate of $1\times10^{-6}$. 
We sample $G=8$ trajectories per query at a temperature of 0.9 and $J=4$ rollouts from each intermediate prefix to estimate prefix values.
The reward weights are set to $\lambda_{\mathrm{fmt}}=0.1$, $\lambda_{\mathrm{gnd}}=0.9$, $\lambda_{\mathrm{TR}}=0.5$, and $\lambda_{\mathrm{multi}}=0.03$.
Each region state contains $M=2$ region latents. 
The auxiliary decoder uses a $4\times4$ grid of positional queries and 8 attention heads, and TAPD performs $R=4$ decoding rounds. Training uses 8 A800 80GB GPUs. More details are in Appendix~\ref{app:training_data}.

\definecolor{refinecolor}{HTML}{D9ECFD}
\definecolor{shiftcolor}{HTML}{DBEFE6}
\definecolor{resetcolor}{HTML}{FAE4D4}
\definecolor{mixturecolor}{HTML}{E7DEF1}

\begin{table*}[!t]
\centering
\caption{Performance of state-of-the-art models on ReferFlow-Bench across four trajectory types. Best results are shown in \textbf{bold}, and the second-best are \underline{underlined}. }
\vspace{-0.2cm}
\label{tab:groundflow_bench}
\setlength{\tabcolsep}{3pt}
\resizebox{\textwidth}{!}{
\begin{tabular}{@{}l@{}c@{\hspace{4pt}}|@{\hspace{4pt}}ccc | ccc | ccc | ccc | ccc}
\toprule

\multirow{2}{*}{\textbf{Model}}
& \multicolumn{1}{c}{\multirow{2}{*}{\textbf{Params.}}}
& \multicolumn{3}{c}{\textbf{Overall}}
& \multicolumn{3}{c}{\textbf{Refinement}}
& \multicolumn{3}{c}{\textbf{Shift}}
& \multicolumn{3}{c}{\textbf{Reset}}
& \multicolumn{3}{c}{\textbf{Mixture}}

\\

&
& \textbf{Q-Acc}
& \textbf{T-Acc}
& \textbf{F-Acc}
& \textbf{Q-Acc}
& \textbf{T-Acc}
& \textbf{F-Acc}
& \textbf{Q-Acc}
& \textbf{T-Acc}
& \textbf{F-Acc}
& \textbf{Q-Acc}
& \textbf{T-Acc}
& \textbf{F-Acc}
& \textbf{Q-Acc}
& \textbf{T-Acc}
& \textbf{F-Acc}
\\
\midrule

\multicolumn{17}{@{}l}{\textit{\textbf{Proprietary MLLMs}}} \\

Gemini 3.1 Pro~\citep{google2026gemini31pro}
& --
& 33.2 & 8.9 & 33.9 & 25.4 & 9.4 & 47.4 & 41.9 & 16.6 & 34.1 & 34.3 & 5.4 & 23.0 & 31.2 & 4.3 & 31.1 \\

GPT-5.6 Sol~\citep{openai2026gpt56}
& --
& 44.9 & 12.7 & 41.5 & 33.4 & 10.9 & 50.6 & 51.8 & \underline{24.0} & 37.8 & 47.0 & 8.4 & 32.6 & 47.5 & 7.4 & 45.1 \\
\midrule
\multicolumn{17}{@{}l}{\textit{\textbf{Open-Source General MLLMs}}} \\

Ovis2.5~\citep{lu2025ovis25technicalreport} & 2B
& 27.6 & 3.2 & 25.4
& 18.2 & 1.8 & 26.8
& 36.7 & 7.2 & 23.2
& 29.6 & 1.2 & 27.9
& 25.9 & 2.4 & 23.6 \\

Qwen3-VL~\citep{bai2025qwen3} & 2B
& 35.6 & 6.3 & 29.6
& 21.4 & 2.7 & 29.9
& 44.3 & 12.3 & 28.3
& 39.9 & 5.8 & 30.9
& 36.9 & 4.3 & 29.2 \\

Qwen3.5~\citep{qwen3.5} & 4B
& 39.1 & 8.7 & 32.5
& 27.1 & 3.3 & 34.7
& 47.4 & 14.9 & 34.9
& 46.2 & 8.2 & 29.6
& 35.6 & 8.3 & 30.6 \\

LLaVA-OneVision-1.5~\citep{an2025llava} & 4B
& 18.4 & 1.9 & 7.8
& 9.4 & 0.4 & 7.2
& 26.8 & 6.4 & 20.0
& 19.4 & 0.6 & 1.7
& 18.1 & 0.0 & 2.4 \\

Youtu-VL~\citep{wei2026youtu} & 4B
& 30.7 & 4.2 & 22.3
& 10.3 & 1.4 & 20.6
& 45.0 & 9.3 & 25.6
& 37.7 & 4.2 & 19.4
& 29.7 & 2.0 & 23.4 \\

MiMo-VL-RL~\citep{yue2025mimo} & 7B
& 29.2 & 3.3 & 22.0
& 18.5 & 1.5 & 20.6
& 38.2 & 7.5 & 27.4
& 30.9 & 2.5 & 20.7
& 29.2 & 1.5 & 19.1 \\

Keye-VL-1.5~\citep{yang2025kwai} & 8B
& 28.1 & 1.9 & 15.5
& 15.1 & 0.8 & 13.2
& 37.2 & 4.5 & 18.0
& 31.3 & 1.5 & 12.7
& 28.7 & 0.9 & 17.9 \\

InternVL3.5~\citep{wang2025internvl3} & 8B
& 29.6 & 5.1 & 23.5
& 18.5 & 4.2 & 21.7
& 39.1 & 9.6 & 26.6
& 29.3 & 2.2 & 21.1
& 31.5 & 4.5 & 24.6 \\

Qwen3.5~\citep{qwen3.5} & 9B
& 41.0 & 9.4 & 35.7
& 28.2 & 3.1 & 35.9
& 51.4 & 15.8 & 36.9
& 48.5 & 8.9 & 31.8
& 35.7 & 9.7 & 38.2 \\

GLM-4.6V-Flash~\citep{vteam2025glm45vglm41vthinkingversatilemultimodal} & 9B
& 42.7 & 11.0 & 38.2
& 26.2 & 5.4 & 42.0
& 52.7 & 20.0 & 36.8
& 46.9 & 8.4 & 31.3
& 45.0 & 10.2 & 42.7 \\

Qwen3.5~\citep{qwen3.5} & 27B
& 47.8 & 12.5 & \underline{44.5}
& 36.2 & 4.8 & 50.6
& 56.7 & 18.7 & 40.6
& 51.7 & 9.7 & 37.1
& 46.6 & \underline{16.6} & \underline{49.5} \\

Qwen3.8~\citep{qwen38} & 27B
& 48.1 & 11.4 & 44.1
& 37.1 & 4.7 & 51.5
& 54.3 & 18.4 & 38.1
& \underline{52.6} & 10.5 & \underline{38.0}
& 48.5 & 12.1 & 48.8 \\

Ovis2.6~\citep{lu2025ovis25technicalreport} & 30B-A3B
& 42.2 & 9.8 & 38.1
& 28.7 & 8.6 & 43.4
& 53.3 & 16.0 & 38.4
& 44.1 & 7.1 & 29.8
& 42.6 & 7.4 & 40.8 \\

InternVL3.5~\citep{wang2025internvl3} & 38B
& 40.2 & 11.9 & 38.1
& 24.5 & 8.3 & 38.7
& 50.0 & 22.8 & 38.4
& 42.8 & 8.2 & 26.8
& 43.3 & 8.1 & 48.4 \\

GLM-4.6V~\citep{vteam2025glm45vglm41vthinkingversatilemultimodal} & 106B-A12B
& 46.4 & 9.6 & 39.5
& 32.3 & 6.0 & 41.1
& 55.6 & 17.2 & 38.0
& 50.1 & 9.5 & 32.7
& 47.6 & 5.5 & 46.0 \\

Qwen3.5~\citep{qwen3.5} & 122B-A10B
& 47.1 & 10.9 & 42.9
& 37.4 & 7.2 & \underline{52.3}
& 56.5 & 19.6 & 40.2
& 50.5 & 10.2 & 33.9
& 43.8 & 6.4 & 45.0 \\

Step-3.7-Flash~\citep{step37flash} & 198B-A11B
& 40.6 & 11.7 & 37.5
& 28.5 & 10.2 & 35.9
& 53.3 & 22.3 & \underline{45.1}
& 45.9 & 9.4 & 32.0
& 34.5 & 5.0 & 36.9 \\

InternVL3.5~\citep{wang2025internvl3} & 241B-A28B
& 42.7 & 11.6 & 38.5
& 23.5 & 6.8 & 35.0
& 52.5 & 20.8 & 40.7
& 49.3 & 10.3 & 34.3
& 45.5 & 8.6 & 44.1 \\

\midrule
\multicolumn{17}{@{}l}{\textit{\textbf{Grounding-Specialized MLLMs}}} \\
DeepEyes~\citep{zheng2026deepeyes} & 7B
& 25.4 & 4.1 & 18.5
& 12.3 & 2.4 & 20.2
& 32.6 & 4.6 & 14.2
& 27.3 & 7.4 & 19.2
& 29.4 & 1.8 & 20.3 \\

RegionReasoner~\citep{sun2026regionreasoner} & 7B
& 28.8 & 4.1 & 19.3
& 14.9 & 0.0 & 7.8
& 36.8 & 8.6 & 23.6
& 38.4 & 4.9 & 22.8
& 25.2 & 2.8 & 22.8 \\

Rex-Omni~\citep{jiang2026detect} & 3B
& \underline{51.2} & \underline{13.4} & 41.1
& \underline{43.8} & \underline{14.2} & 50.6
& \underline{58.1} & 20.6 & 42.8
& 48.6 & \underline{12.1} & 36.8
& \underline{54.3} & 6.6 & 34.0 \\

\midrule
\textbf{ReferFlow} & \textbf{2B}
& \textbf{63.9} & \textbf{28.4} & \textbf{63.0}
& \textbf{56.3} & \textbf{22.6} & \textbf{65.8}
& \textbf{69.8} & \textbf{36.0} & \textbf{62.2}
& \textbf{66.1} & \textbf{29.7} & \textbf{63.4}
& \textbf{63.4} & \textbf{25.2} & \textbf{60.4} \\

\bottomrule
\end{tabular}
}
\end{table*}

\begin{table*}[!t]
    \centering
    \caption{Comparisons on general referring expression comprehension benchmarks. Best results are shown in \textbf{bold}, and the second-best are \underline{underlined}. “--” indicates unavailable or unreported results.}
    \vspace{-0.2cm}
    \label{tab:main_results}
    \resizebox{\textwidth}{!}{
    \begin{tabular}{lc@{\hspace{19pt}}ccc@{\hspace{19pt}}ccc@{\hspace{19pt}}cc}
    \toprule
    \textbf{Model} & \textbf{Params.} &
    \textbf{RefCOCO} & \textbf{RefCOCO+} & \textbf{RefCOCOg} &
    \textbf{Ref-L4} & \textbf{Ref-Adv} & \textbf{GroundingME} &
    \textbf{HumanRef} & \textbf{PR-Bench} \\
    \midrule
    
    \multicolumn{10}{l}{\textit{\textbf{General MLLMs}}} \\
    
    Qwen3-VL~\citep{bai2025qwen3}
    & 2B
    & 87.4 & 78.1 & 84.9 & 85.4 & 42.5 & 26.6 & 58.9 & 52.8 \\
    
    Qwen3.5~\citep{qwen3.5}
    & 4B
    & 90.1 & 83.5 & 88.4 & 88.7 & 59.3 & 39.6 & 69.0 & 62.4 \\
    
    Youtu-VL~\citep{wei2026youtu}
    & 4B
    & 93.2 & \underline{89.8} & \underline{92.6} & 83.3 & 48.5 & 14.1 & 55.3 & 53.2 \\
    
    InternVL3.5~\citep{wang2025internvl3}
    & 8B
    & 91.9 & 87.6 & 89.5 & 80.9 & 49.7 & 11.8 & 40.0 & 41.5 \\
    
    GLM-4.6V-Flash~\citep{vteam2025glm45vglm41vthinkingversatilemultimodal}
    & 9B
    & 88.1 & 80.0 & 88.7 & 72.7 & 58.0 & 29.1 & 75.2 & 58.9 \\
    
    Qwen3.5~\citep{qwen3.5}
    & 27B
    & 92.5 & 89.6 & 90.6 & 90.2 & \underline{66.7} & \underline{42.2} & 71.5 & \underline{65.3} \\
    
    InternVL3.5~\citep{wang2025internvl3}
    & 38B
    & 90.4 & 87.4 & 89.8 & 85.5 & 57.6 & 15.7 & 51.1 & 48.5 \\
    
    \midrule
    \multicolumn{10}{l}{\textbf{\textit{Grounding-Specialized MLLMs}}} \\
    
    Rex-Thinker~\citep{jiang2026rex}
    & 7B
    & 85.9 & 82.9 & 83.3 & 70.2 & 42.8 & 34.7 & \underline{83.5} & 44.0 \\
    
    RegionReasoner~\citep{sun2026regionreasoner}
    & 7B
    & 86.4 & 88.6 & 88.4 & 66.8 & 50.6 & 36.1 & 71.6 & 48.8 \\
    
    DeepEyes~\citep{zheng2026deepeyes}
    & 7B
    & 89.8 & 83.6 & 86.7 & 85.2 & 45.3 & 13.2 & 50.3 & 45.7 \\
    
    VGent~\citep{kang2026vgent}
    & 7B
    & 92.3 & 87.9 & 90.3 & -- & -- & -- & -- & -- \\
    
    Rex-Omni~\citep{jiang2026detect}
    & 3B
    & 86.3 & 78.6 & 86.7 & 78.0 & 45.1 & 28.8 & 74.9 & 53.7 \\
    
    WeDetect-Ref~\citep{fu2026wedetect}
    & 2B
    & \underline{94.2} & 88.1 & 92.1 & \underline{93.9} & 52.5 & 29.3 & 79.8 & 55.7 \\
    
    LocateAnything~\citep{wang2026locateanything}
    & 3B
    & 92.2 & 86.1 & 88.9 & 87.9 & 54.1 & 23.0 & 68.8 & 57.4 \\
    
    \midrule
    % \rowcolor{gray!10}
    \textbf{ReferFlow}
    & 2B
    & \textbf{94.6} & \textbf{90.6} & \textbf{92.9} & \textbf{94.3}
    & \textbf{67.7} & \textbf{43.0} & \textbf{83.9} & \textbf{66.1} \\
    
    \bottomrule
    \end{tabular}
    }
    \end{table*}

\subsection{Main Results}

\noindent \textbf{Trajectory-Level REC.}
Table~\ref{tab:groundflow_bench} compares MLLMs on ReferFlow-Bench. Existing models struggle to localize target regions consistently throughout the trajectory. For example, Qwen3.8-27B and GPT-5.6 Sol achieve only 11.4\% and 12.7\% T-Acc, respectively.
In Refinement, additional constraints narrow the target set, but localizing multiple target regions at intermediate steps remains difficult. Even Rex-Omni, which leads the baselines on both Q-Acc and T-Acc for this type, reaches only 43.8\% and 14.2\%, respectively.
In Shift, models show a basic ability to use previously localized regions as references for locating related targets.
Reset instead requires localizing new targets independently while avoiding interference from prior grounding, with even the strongest baseline reaching only 12.1\% T-Acc.
Mixture combines multiple transition patterns and requires models to adapt to changing dependencies on prior grounding, with the strongest baseline reaching 16.6\% T-Acc.
In contrast, ReferFlow achieves 63.9\% Q-Acc, 28.4\% T-Acc, and 63.0\% F-Acc, exceeding the strongest baseline for each metric by 12.7\%, 15.0\%, and 18.5\% respectively.

\noindent \textbf{General REC.}
ReferFlow achieves strong performance across REC benchmarks with varying query complexity. As shown in Tab.~\ref{tab:main_results}, ReferFlow achieves leading results on RefCOCO/+/g and Ref-L4, covering queries from short phrases to detailed descriptions.
Furthermore, ReferFlow adaptively decomposes complex queries involving multiple visual references into grounding steps, using intermediate regions to support target localization.
On Ref-Adv, ReferFlow achieves 67.7\%, showing discrimination of visually similar distractors.
For tasks requiring fine-grained understanding and compositional reasoning, it surpasses LocateAnything by +20.0\%, +15.1\%, and +8.7\% on GroundingME, HumanRef, and PR-Bench, respectively, despite its smaller model size.

\subsection{Analysis and Ablations}

\begin{figure}[!t]
    \centering
        \includegraphics[width=\linewidth]{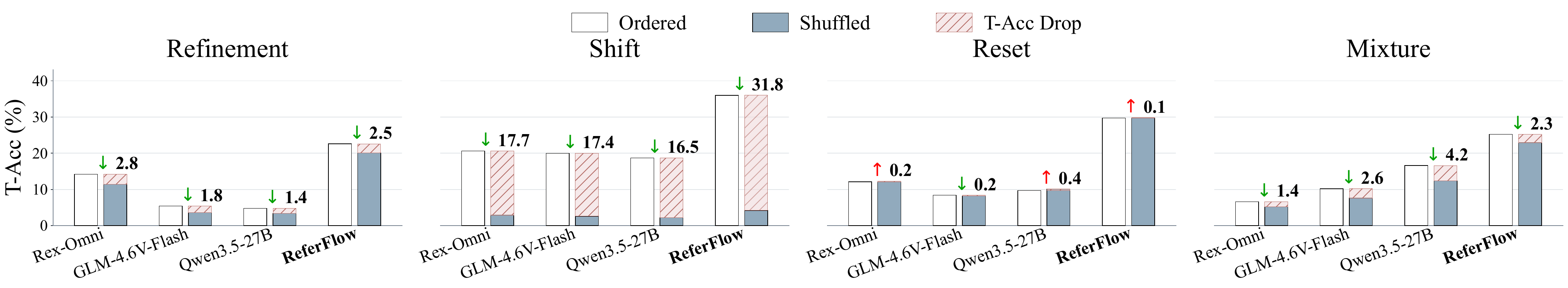}
    \caption{Comparison of T-Acc  between ordered and shuffled grounding steps.}
    \label{fig:order_shuffle}
\end{figure}

\noindent \textbf{Effects of Grounding-Step Order.}
We compare T-Acc before and after shuffling grounding steps while preserving each step’s target-region annotations (Fig.~\ref{fig:order_shuffle}).
For Shift, T-Acc drops to nearly zero for most models, consistent with its reliance on region references established in earlier steps.
Performance on Reset remains largely unchanged, as its targets are specified independently of prior grounding.
For Refinement, performance declines when shuffling disrupts the progression from broader to narrower target sets.
Mixture similarly degrades as shuffling disrupts dependencies across its combined transition types.
These results highlight type-specific cross-step dependencies.

\begin{figure}[!t]
    \centering
     \includegraphics[width=\linewidth]{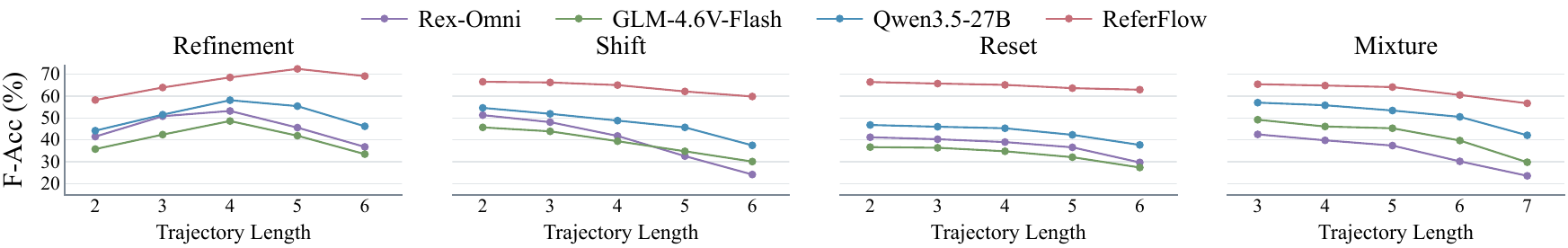}
    \caption{Comparison of F-Acc  across different trajectory lengths.}
    \label{fig:trajectory_length}
\end{figure}

\begin{figure}[!t]
    \centering
        \includegraphics[width=\linewidth]{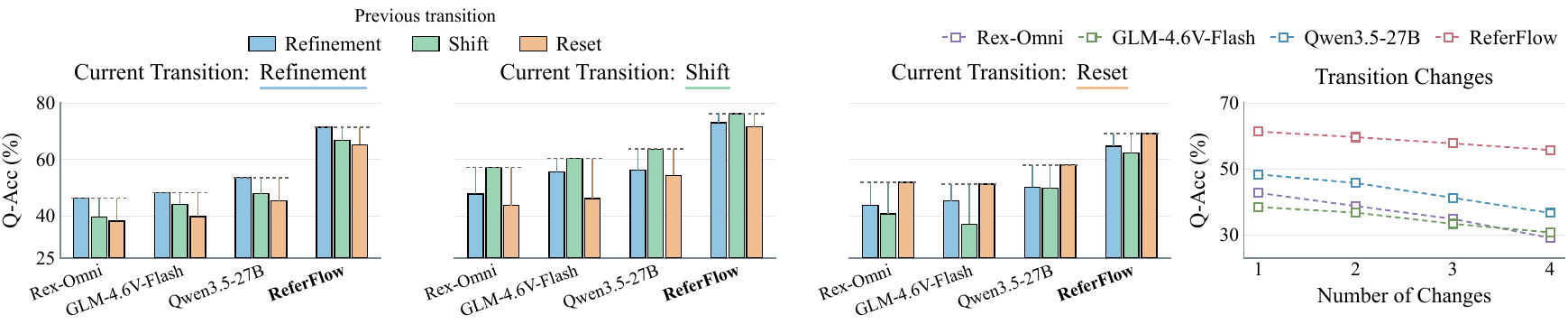}
    \caption{Comparison of Q-Acc across different types and numbers of transition changes.}
    \label{fig:bmk2_transition}
\end{figure}

\noindent \textbf{Effects of Trajectory Length.}
Fig.~\ref{fig:trajectory_length} reports F-Acc by trajectory length.
(1) For Refinement, F-Acc initially increases as constraints narrow the target set, but declines on longer trajectories, suggesting difficulty integrating more constraints.
ReferFlow peaks at longer trajectory lengths than the baselines, indicating that it benefits from additional refinement steps.
(2) For Reset, F-Acc declines more sharply on longer trajectories, consistent with increasing interference from prior grounding during independent target localization.
 (3) F-Acc on Shift and Mixture decreases substantially with length, suggesting difficulty handling repeated target transitions and changing cross-step dependencies.

\noindent \textbf{Effects of Transition Changes.}
We evaluate fixed-length Mixture trajectories with different transition patterns and numbers of type changes in Fig.~\ref{fig:bmk2_transition}.
Q-Acc consistently decreases when transition types change, indicating difficulty in switching between grounding behaviors.
Switches between Shift and Reset are particularly challenging, as models must alternate between using prior regions as references and localizing new targets independently.
Q-Acc declines further with more frequent type changes, suggesting that repeated switching hinders consistent grounding.

\begin{table*}[t]
    \centering
    \caption{
        Ablation studies of SRLR under different spatial and visual supervision settings.
    }
    \vspace{-0.2em}
    \label{tab:srlr_ablation}
    \footnotesize
    \renewcommand{\arraystretch}{0.92}
    \resizebox{\linewidth}{!}{
    % Share the single-line group heading width evenly across its five columns.
    \settowidth{\dimen0}{\textbf{ReferFlow-Bench Q-Acc by Trajectory Length}}
    \advance\dimen0 by -8\tabcolsep
    \divide\dimen0 by 5
    \advance\dimen0 by 0.1pt
    \begin{tabular}{lcc|ccccc|ccccc}
        \toprule
        \multirow{2}{*}{\textbf{Setting}}
        & \multicolumn{2}{c|}{\textbf{Supervision}}
        & \multirow{2}{*}{\textbf{RefCOCOg}}
        & \multirow{2}{*}{\textbf{GroundingME}}
        & \multicolumn{3}{c|}{\textbf{ReferFlow-Bench}}
        & \multicolumn{5}{c}{\textbf{ReferFlow-Bench Q-Acc by Trajectory Length}} \\
        \cmidrule(lr){2-3}
        \cmidrule(lr){6-8}
        \cmidrule(lr){9-13}
        & Spatial & Visual
        & & 
        & Q-Acc & T-Acc & F-Acc
        & \makebox[\dimen0]{2} & \makebox[\dimen0]{3} & \makebox[\dimen0]{4} & \makebox[\dimen0]{5} & \makebox[\dimen0]{6+} \\
        \midrule

        Text
        & -- & --
        & 87.9 & 32.4
        & 52.6 & 17.9 & 48.2
        & 58.8 & 54.7 & 50.1 & 44.0 & 41.2 \\

        Region Latent
        & $\times$ & $\times$
        & 88.5 & 32.8
        & 53.0 & 18.1 & 48.7
        & 59.6 & 55.4 & 49.8 & 43.9 & 40.8 \\

        Region Latent
        & $\checkmark$ & $\times$
        & 89.7 & 34.2
        & 54.4 & 19.3 & 50.4
        & 60.8 & 56.6 & 52.0 & 45.3 & 42.0 \\

        Region Latent
        & $\times$ & $\checkmark$
        & 89.1 & 35.8
        & 55.4 & 20.0 & 52.1
        & 60.3 & 56.5 & 53.6 & 49.1 & 46.4 \\

        SRLR
        & $\checkmark$ & $\checkmark$
        & 90.5 & 37.4
        & 56.9 & 21.3 & 53.9
        & 61.8 & 57.8 & 55.0 & 51.0 & 48.9 \\

        \bottomrule
    \end{tabular}
    }
\end{table*}

% Side-by-side module and training ablations.

\begin{table}
    \centering
    \caption{
    Ablations of (a) TAPD parallelization and coordinate recovery and (b) grounding modes under SFT and trajectory-level reward components.
    In (a), Intra-Box and Inter-Box denote parallel decoding within and across boxes, respectively.
    Grouping recovers each three-digit coordinate as an atomic unit, while Geom. enforces geometric constraints.
}
    \label{tab:unified_ablation}
    \label{tab:reasoning_decoding_ablation}
    \label{tab:adaptive_trajectory_ablation}

    \scriptsize
    \setlength{\tabcolsep}{2.1pt}
    \renewcommand{\arraystretch}{1.0}
    \ifdefined\tapdAblationBox\else\newsavebox{\tapdAblationBox}\fi
    \ifdefined\trajectoryAblationBox\else\newsavebox{\trajectoryAblationBox}\fi
    \sbox{\tapdAblationBox}{%
        \settowidth{\dimen0}{\textbf{ReferFlow-Bench}}
        \advance\dimen0 by -2\tabcolsep
        \divide\dimen0 by 2
        \advance\dimen0 by 0.1pt
        \begin{tabular}{lcc|cccc|c}
            \toprule
    
            \multicolumn{3}{l|}{\multirow[c]{2}{*}{\textbf{Setting}}}
            & \multirow[c]{2}{*}{\textbf{RefCOCOg}}
            & \multirow[c]{2}{*}{\textbf{GroundingME}}
            & \multicolumn{2}{c|}{\textbf{ReferFlow-Bench}}
            & \multirow[c]{2}{*}{\shortstack[c]{\textbf{Valid Box}\\\textbf{Rate}}} \\
            
            \cline{6-7}
            
            & &
            & &
            % Lower the labels within the fixed-height row to balance rule gaps.
            & \makebox[\dimen0]{\raisebox{-1.35pt}[0pt][0pt]{Q-Acc}}
            & \makebox[\dimen0]{\raisebox{-1.35pt}[0pt][0pt]{T-Acc}}
            & \\
    
            \midrule
    
            % =====================================================
            % Parallelization
            % =====================================================
            \multicolumn{1}{l}{\textit{Parallel Decoding}}
            & \multicolumn{1}{c}{Intra-Box}
            & \multicolumn{1}{c|}{Inter-Box}
            & & & & & \\
    
            NTP
            & $\times$ & $\times$
            & 90.5 & 37.4 & 56.9 & 21.3 & 95.7 \\

            NTP
            & $\times$ & $\checkmark$
            & 90.8 & 37.9 & 58.1 & 22.5 & 96.3 \\
    
            MTP
            & $\checkmark$ & $\times$
            & 87.9 & 33.4 & 50.3 & 15.0 & 88.3 \\
    
            MTP
            & $\checkmark$ & $\checkmark$
            & 88.6 & 34.9 & 53.8 & 18.2 & 90.4 \\
    
            \midrule
    
            % =====================================================
            % Recovery Design
            % =====================================================
            \multicolumn{1}{l}{\textit{Recovery Design}}
            & \multicolumn{1}{c}{Grouping}
            & \multicolumn{1}{c|}{Geom.}
            & & & & & \\
    
            Token-wise
            & $\times$ & $\times$
            & 91.1 & 38.2 & 58.3 & 22.7 & 96.8 \\
    
            Coordinate-wise
            & $\checkmark$ & $\times$
            & 91.6 & 38.6 & 59.7 & 24.3 & 98.2 \\
    
            TAPD
            & $\checkmark$ & $\checkmark$
            & 92.0 & 39.1 & 60.8 & 25.6 & 100.0 \\
    
            \bottomrule
        \end{tabular}%
    }
    \sbox{\trajectoryAblationBox}{%
        \settowidth{\dimen2}{Adaptive Trajectory }
        \advance\dimen2 by -2\tabcolsep
        \divide\dimen2 by 2
        \settowidth{\dimen0}{\textbf{ReferFlow-Bench}}
        \advance\dimen0 by -2\tabcolsep
        \divide\dimen0 by 2
        \advance\dimen0 by 0.1pt
        \begin{tabular}{cc|cccc}
            \toprule

            \multicolumn{2}{l|}{\multirow[c]{2}{*}{\textbf{Setting}}}
            & \multirow[c]{2}{*}{\textbf{RefCOCOg}}
            & \multirow[c]{2}{*}{\textbf{GroundingME}}
            & \multicolumn{2}{c}{\textbf{ReferFlow-Bench}} \\
            \cline{5-6}
            & & &
            & \makebox[\dimen0]{\raisebox{-1.35pt}[0pt][0pt]{Q-Acc}}
            & \makebox[\dimen0]{\raisebox{-1.35pt}[0pt][0pt]{T-Acc}} \\
            \midrule

            % ---------------- SFT ----------------
            \multicolumn{2}{l|}{\textit{Grounding Modes}} & & & & \\

            \multicolumn{2}{l|}{Direct Grounding}
            & 91.1 & 35.2 & -- & -- \\

            \multicolumn{2}{l|}{Trajectory Only  }
            & 88.7 & 37.9 & 60.3 & 25.0 \\

            \multicolumn{2}{l|}{Adaptive Trajectory  }
            & 92.0 & 39.1 & 60.8 & 25.6 \\

            \midrule

            % ---------------- RL ----------------
            \multicolumn{2}{l|}{\textit{Reward Design}} & & & & \\

            \multicolumn{1}{c}{\makebox[\dimen2]{$R_{\mathrm{TR}}$}}
            & \multicolumn{1}{c|}{\makebox[\dimen2]{$R_{\mathrm{multi}}$}}
            & & & & \\

            $\times$
            & $\times$
            & 92.7 & 40.6 & 61.8 & 26.2 \\

            $\checkmark$
            & $\times$
            & 92.8 & 42.1 & 63.4 & 28.1 \\

            $\checkmark$
            & $\checkmark$
            & 92.9 & 43.0 & 63.9 & 28.4 \\

            \bottomrule
        \end{tabular}%
    }
    \resizebox{\linewidth}{!}{%
        \begin{minipage}[t]{\wd\tapdAblationBox}
            \vspace{0pt}
            \centering
            {\small\bfseries (a) TAPD Design\par}
            \vspace{4pt}
            \usebox{\tapdAblationBox}
        \end{minipage}%
        \hspace{14.5pt}%
        \begin{minipage}[t]{\wd\trajectoryAblationBox}
            \vspace{0pt}
            \centering
            {\small\bfseries (b) Adaptive Trajectory Generation\par}
            \vspace{4pt}
            \usebox{\trajectoryAblationBox}
        \end{minipage}%
    }

\end{table}

\noindent \textbf{Effects of SRLR.}
Tab.~\ref{tab:srlr_ablation} evaluates the region-state representation and its visual and spatial supervision. Region latents without explicit supervision yield only marginal gains and underperform text-only reasoning on trajectories of lengths 5 and 6+. Spatial supervision provides consistent gains across lengths and performs better on shorter trajectories, whereas visual supervision yields larger gains on longer trajectories. These gains increase with trajectory length, highlighting the importance of retaining target-specific visual content across grounding steps.
 Combining both supervision achieves the best performance across all metrics, demonstrating their complementary roles.

\noindent \textbf{Effects of TAPD.} 
Tab.~\ref{tab:reasoning_decoding_ablation}(a) compares decoding strategies based on the same SRLR setting.
Across-box parallel decoding improves both NTP and MTP, with larger gains on ReferFlow-Bench.
However, one-shot within-box MTP underperforms NTP across all metrics and sharply reduces box validity, indicating difficulty maintaining coordinate consistency in a single pass.
Iterative token-wise recovery addresses this limitation by conditioning each round on previously recovered high-confidence digits, outperforming NTP with across-box parallel decoding.
Coordinate-wise recovery yields further gains by recovering each three-digit coordinate as an atomic unit.
Geometry-constrained recovery restricts candidates to feasible intervals, yielding the best overall performance.

\noindent \textbf{Effects of Adaptive Trajectory Generation.}
Tab.~\ref{tab:adaptive_trajectory_ablation}(b) compares grounding modes and rewards.
Trajectory Only outperforms Direct Grounding on complex queries but slightly reduces RefCOCOg accuracy.
Adaptive trajectory generation allows the model to directly localize targets or generate intermediate grounding steps as needed, maintaining strong performance on both simple and complex queries.
Regarding reward design, the trajectory reward $R_{\mathrm{TR}}$ encourages intermediate prefixes that support final localization, while the multi-step reward $R_{\mathrm{multi}}$ encourages successful multi-step grounding. Consequently, the full reward formulation achieves the best overall performance.

\section{Conclusion}

We formulate trajectory-level REC and build ReferFlow-Bench for evaluation.
Evaluating over 20 MLLMs reveals difficulty in consistently grounding intermediate visual references.
ReferFlow combines SRLR for object-indexed region states with TAPD for parallel geometry-constrained box decoding.
Grounding-Aware GRPO promotes consistent grounding and adaptive trajectory generation.
ReferFlow achieves leading performance on general REC benchmarks and ReferFlow-Bench.

\newpage
\bibliographystyle{plainnat}
\bibliography{paper}

\clearpage
\appendix
\section{Background and Preliminaries}

\subsection{Latent Visual Reasoning}

\noindent\textbf{Latent Visual Reasoning.}
Inspired by methods such as Coconut~\citep{hao2024training}, which feeds hidden states back as continuous input embeddings, recent work extends latent reasoning to MLLMs to overcome the limitations of text-only reasoning.
LVR~\citep{li2026latent} aligns latent states with query-relevant visual embeddings to support visual question answering.
ILVR~\citep{dong2026interleaved} interleaves textual reasoning with visual latents supervised by selectively distilled features from intermediate images.
Future-L1~\citep{jiang2026imagine} extends interleaved latent reasoning to video event prediction, using future-frame embeddings to supervise intermediate visual states.
ProLaViT~\citep{li2026prolavit} learns structured, progressive latent visual thoughts by reconstructing features of synthesized intermediate images.
These methods employ continuous latent representations to preserve visual semantics and facilitate multimodal reasoning toward a final prediction. In contrast, ReferFlow targets trajectory-level REC by learning object-indexed region states. These states retain spatial information for localization and preserve fine-grained visual content as evidence across grounding steps. This method maintains visual evidence in a continuous form, avoiding the information loss inherent in reducing it to discrete textual descriptions. Furthermore, each region state corresponds to an explicit bounding box prediction, thereby integrating both intermediate and final localization directly into the task output.

\subsection{Coordinate Decoding}
Given an image \(I\) and a textual query \(q\), MLLM-based grounding methods typically discretize each bounding box \(\mathbf{b}=(x_{\min},y_{\min},x_{\max},y_{\max})\) into a coordinate block \(\mathcal{B}=(b_1,\ldots,b_{L_b})\) of \(L_b\) tokens, which is then integrated with textual outputs in a unified sequence \(\mathbf{x}=(x_1,\ldots,x_L)\).

\noindent\textbf{Next-Token Coordinate Decoding.}
Next-token prediction (NTP)~\citep{brown2020language} models the output sequence as:
\begin{equation}
\log p_\theta(\mathbf{x}\mid I,q)
=
\sum_{\ell=1}^{L}
\log p_\theta\left(x_\ell\mid I,q,\mathbf{x}_{<\ell}\right),
\end{equation}
where $\mathbf{x}_{<\ell}$ denotes the preceding tokens.
Each coordinate token is conditioned on previously generated text and coordinates.
Thus, coordinate decoding is sequential within and across boxes, requiring one decoding step per coordinate token.

\noindent\textbf{Multi-Token Coordinate Decoding.}
Multi-token prediction (MTP)~\citep{wang2026locateanything} organizes each bounding box into a fixed-length coordinate block and jointly predicts its coordinate tokens. For the \(k\)-th coordinate block \(\mathcal{B}_k=(b_{k,1},\ldots,b_{k,L_b})\), the conditional distribution is:
\begin{equation}
p_\theta(\mathcal{B}_k\mid C_k)
=
\prod_{\ell=1}^{L_b}
p_\theta\left(b_{k,\ell}\mid C_k\right),
\end{equation}
where \(C_k\) denotes the context of the \(k\)-th coordinate block. MTP removes autoregressive dependencies within each block. 
However, multiple  blocks are still generated sequentially, resulting in cross-block dependencies, and each block is predicted in a single pass without iterative recovery.

\noindent\textbf{Iterative Masked Decoding.}
Masked diffusion models such as LLaDA~\citep{nie2026large} have shown effectiveness  in general language generation, producing sequences by progressively recovering masked tokens.
Let $C$ denote the fixed conditioning context, $\mathbf{x}^{(r)}$ the partially masked sequence at round $r$, and $\mathcal{M}_r$ its masked positions.
In the greedy confidence-based variant, each masked position is assigned a prediction and confidence:
\begin{equation}
\begin{aligned}
\widehat{x}_{\ell}^{(r)}
&=
\operatorname*{arg\,max}_{v\in\mathcal{V}}
p_\theta\left(x_\ell=v\mid C,\mathbf{x}^{(r)}\right),\\
c_{\ell}^{(r)}
&=
p_\theta\left(x_\ell=\widehat{x}_{\ell}^{(r)}
\mid C,\mathbf{x}^{(r)}\right),
\qquad \ell\in\mathcal{M}_r,
\end{aligned}
\end{equation}
where $\mathcal{V}$ is the token vocabulary.
Starting from a fully masked sequence, the model recovers a scheduled number of high-confidence tokens at each round while keeping previously recovered tokens fixed.
Similarly, SDAR~\citep{cheng2026sdar} adapts pretrained autoregressive language models to this paradigm through masked-token training, iteratively recovering tokens within each block while preserving autoregressive dependencies across blocks. However, extending masked decoding to general language generation requires extra training to align pretrained models with a different prediction objective and attention mechanism.
Coordinate prediction is naturally suited for parallel decoding, as bounding boxes have a fixed structure and do not require a sequential, left-to-right generation order. For example, DETR~\citep{carion2020end} uses learned object queries to attend to image features and predict bounding boxes in parallel. 
However, this method does not explicitly generate intermediate grounding steps for complex queries. MLLMs can generate these steps autoregressively, leaving coordinate recovery to parallel decoding. Yet, standard token-wise masked decoding may recover the digits of a coordinate in different rounds without enforcing box geometry. To solve this, ReferFlow decouples text-latent trajectory generation from coordinate decoding and recovers all boxes in parallel through coordinate-wise, geometry-constrained refinement. This reduces within-box sequential dependencies and eliminates cross-box autoregression.

\section{Details of ReferFlow-Bench}\label{bmk_details}

\subsection{Annotation Pipeline} \label{annotation_pipeline}
\begin{figure}[!htbp]
  \centering
  \includegraphics[width=\linewidth]{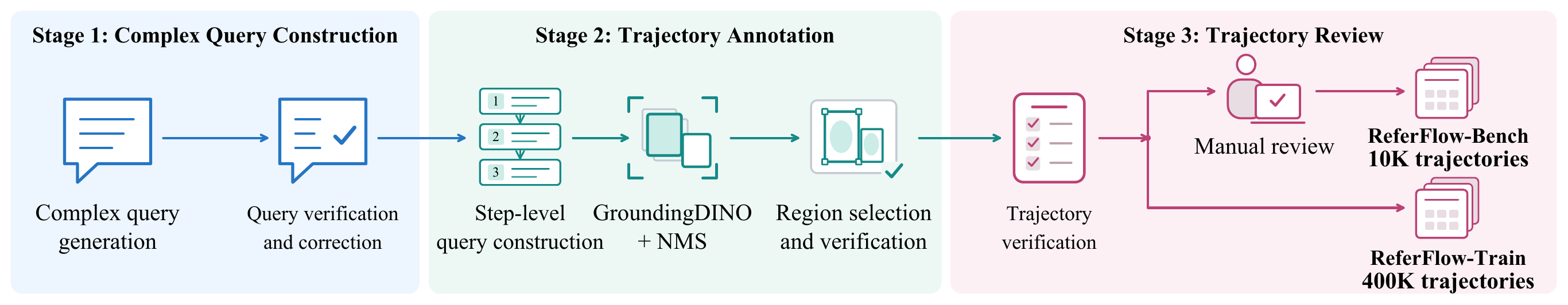}
  \caption{Overview of the annotation pipeline for ReferFlow-Bench and ReferFlow-Train.}
  \label{fig:annotation_pipeline}
\end{figure}

We select object-dense images containing multiple distinguishable instances from Open Images v7~\citep{DBLP:journals/ijcv/KuznetsovaRAUKP20}, without using its original annotations in dataset construction.
As illustrated in Fig.~\ref{fig:annotation_pipeline}, our annotation pipeline consists of three stages: (1) Complex Query Construction, (2) Trajectory Annotation, and (3) Trajectory Review. The first stage generates and verifies complex queries based on image content. The second constructs ordered grounding steps for each query and annotates their target regions. The third automatically verifies complete trajectories and splits them by source image into benchmark and training sets, with two rounds of manual review for the benchmark split. Qwen3.5-27B performs all MLLM-based generation and verification.

\noindent\textbf{Complex Query Construction.}
For each selected image, we prompt the MLLM to generate a complex query based on visible objects, attributes, and relationships. 
Task-specific prompts prescribe the intended trajectory type (Refinement, Shift, Reset, or Mixture), the trajectory length, and the number of targets.
The MLLM then checks each query for factual accuracy, verifying that the mentioned objects are present in the image and that their attributes and relationships are correctly described.
Queries with factual errors are revised without changing their intended final targets.

\noindent\textbf{Trajectory Annotation.}
For each verified complex query, we construct an ordered sequence of grounding steps, each formulated as a step-level query. 
For each step-level query, we use GroundingDINO~\citep{liu2024grounding} to generate candidate bounding boxes and apply non-maximum suppression (NMS) to remove redundant boxes.
Given the full image and the regions from preceding steps, the MLLM checks whether the candidate regions correctly and completely identify the targets of the current step-level query. Trajectories containing steps that fail this check are discarded.

\noindent\textbf{Trajectory Review.}
The MLLM reviews the complete trajectory to ensure that the ordered steps and their dependencies preserve the semantics of the original complex query and accurately locate the intended final targets. Trajectories with unresolved inconsistencies are discarded.
We split the trajectories by source image into a benchmark set and a training set.
For ReferFlow-Bench, three annotators conduct two rounds of manual review, accepting or rejecting each trajectory based on the accuracy of its query, bounding boxes, and grounding steps.
They achieve an initial inter-annotator agreement of 92.8\%, and 10K trajectories are retained after review.
The training split constitutes ReferFlow-Train, which contains 400K automatically verified trajectories.
An additional audit of 2,000 ReferFlow-Train trajectories finds that 89.1\% satisfy the same criteria without modification.

\subsection{Statistics of ReferFlow-Bench}
\begin{figure}[!htbp]
  \centering
  \includegraphics[width=\linewidth]{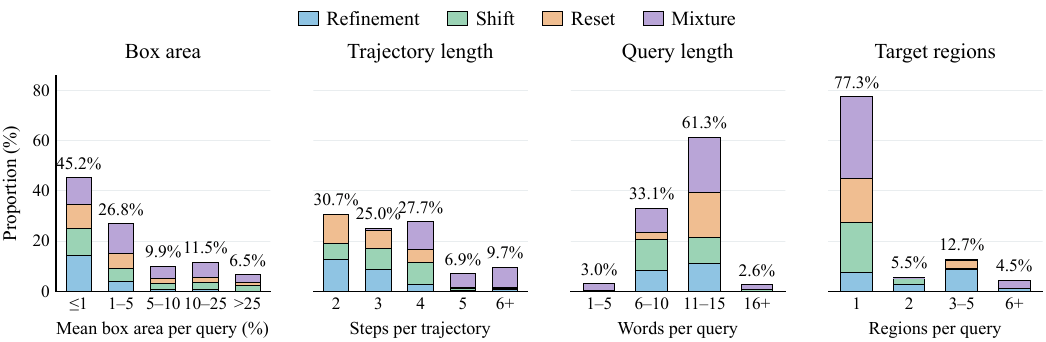}
  \caption{\textbf{Statistics of ReferFlow-Bench.}  Distributions of (a) mean box area per query, (b) trajectory length, (c) query length, and (d) the number of target regions per query. Query-level statistics are computed over step-level queries.}
  \label{fig:supp_bench_statistics}
\end{figure}

ReferFlow-Bench comprises 10,000 trajectories with an average image resolution of  990$\times$769. The dataset features a mean target-region area ratio of 5.76\% and an average step-level query length of 11 words. As illustrated in Fig.~\ref{fig:supp_bench_statistics}, 45.2\% of the queries exhibit a mean relative bounding box area of at most 1\%, while 72.1\% have a mean relative area of at most 5\%. Trajectory lengths range from 2 to 8 steps. Specifically, 83.4\% consist of 2 to 4 steps, and 16.6\% comprise five or more. In terms of query length, 61.3\% of the queries contain 11 to 15 words, followed by 33.1\% that contain 6 to 10 words. Regarding spatial complexity, single-region queries account for 77.3\% of all step-level queries, whereas the remaining 22.7\% require multiple regions, with 4.5\% involving six or more.

\subsection{Details of the Evaluation Protocol}
\label{app:bench_evaluation_protocol}
ReferFlow-Bench employs a single-turn evaluation protocol. Each model receives an image and the complete ordered sequence of grounding steps derived from a single complex query. The model is required to output the target bounding boxes for each step in the specified order within a single response. Crucially, the ground-truth bounding boxes and the number of target objects at each step are not provided to the model. For example, in the Shift task shown in Fig.~\ref{fig:bmk}, the evaluation prompt is \textit{``Locate the man in black at the chessboard, then the woman seated beside him, and finally the scoresheet she uses to record the moves. Return all target boxes for each step in the specified order in a single response.''}
We compare the set of predicted bounding boxes for the $t$-th step with the corresponding ground-truth set. Missing or unparseable predictions are treated as incorrect. Any predicted steps exceeding the specified trajectory length $T$ are excluded from the Q-Acc and F-Acc calculations. A step is deemed correct only if the predicted and ground-truth sets contain an identical number of boxes and can be matched one-to-one such that every pair achieves an $\mathrm{IoU} \geq 0.5$, irrespective of the box order. Q-Acc averages the proportion of correctly grounded steps across trajectories, utilizing the specified length $T$ as the denominator for each trajectory. T-Acc measures the proportion of trajectories in which every step is correct without any extra predicted steps. Finally, F-Acc measures accuracy at the final step $T$.

\section{Details of ReferFlow}\label{model_details}
\subsection{Details of Training Data and Trajectory Supervision}
\label{app:training_data}

We construct our training data by combining ReferFlow-Train with existing REC datasets. For supervised fine-tuning, we compile 1.6M samples, comprising 400K annotated grounding trajectories from ReferFlow-Train and 1.2M REC samples randomly sampled from the training sets of RefCOCO/+/g~\citep{kazemzadeh2014referitgame,yu2016modeling}, Rexverse-2M~\citep{jiang2024chatrextamingmultimodalllm}, Visual Genome~\citep{krishna2017visual}, and HumanRef-CoT-45k~\citep{jiang2026rex}. 

Standard REC samples take a referring expression (e.g., \textit{``Locate the red cup on the table''}) as input and are formulated as single-step trajectories with $T=1$. 
We formulate the ReferFlow-Train data in two distinct formats, with each trajectory randomly assigned to only one format.
For trajectory-level REC, the input provides ordered grounding steps (e.g., \textit{``Locate the man in a blue shirt, then the cup in his left hand''}), and the model is trained to localize the target at each step. 
For trajectory generation, the input is the original complex query.
The target response contains an ordered sequence of annotated step-level queries paired with their corresponding bounding boxes, supervising model to decompose the query into grounding steps and localize the targets at each step.

\begin{wraptable}{r}{0.35\linewidth}
    \vspace{-\intextsep}
    \centering
    \scriptsize
    \captionsetup{font=footnotesize}
    \caption{
    Composition of training data.
    }
    \label{tab:training_data}
    \setlength{\tabcolsep}{3pt}
    \renewcommand{\arraystretch}{1.08}
    \begin{tabular*}{\linewidth}{@{}l@{\extracolsep{\fill}}cc@{}}
        \toprule
        \textbf{Dataset} & \textbf{SFT} & \textbf{RL} \\
        \midrule
        ReferFlow-Train & 400K & 3K \\
        RefCOCO/+/g      & 320K & 1K \\
        Rexverse-2M      & 400K & 5K \\
        Visual Genome    & 450K & 2K \\
        HumanRef-CoT-45k         & 30K  & 2K \\
        \midrule
        \textbf{Total}   & \textbf{1.6M} & \textbf{13K} \\
        \bottomrule
    \end{tabular*}
\end{wraptable}

For Grounding-Aware GRPO, we select 13K challenging samples from the SFT dataset.
We first perform rollouts using the SFT checkpoint and select samples based on final-target localization, prioritizing those with incorrect predictions.
The selected data contain samples from both ReferFlow-Train and existing REC datasets, as summarized in Tab.~\ref{tab:training_data}.
For REC, the model receives an image and a referring expression, and adaptively performs direct or trajectory-based grounding.
For trajectory-level REC, the model receives the image and an ordered sequence of grounding steps.
Grounding rewards use only final-target annotations for REC and are averaged over the steps for trajectory-level REC.

\subsection{More Implementation Details}
\label{more_details}

\noindent\textbf{Region State Supervision.}
For visual supervision, we encode reference features from the ground-truth target region.
For each normalized ground-truth bounding box $\mathbf{b}^*=(x_{\min},y_{\min},x_{\max},y_{\max})$, we map it to the original image coordinates, crop the corresponding region, and extract region features using the same processor as the backbone and a frozen vision encoder.
The crop boundaries are clipped to the image range, with the minimum crop width and height set to 10 pixels to avoid invalid regions after conversion to integer pixel coordinates.
Given the resulting region feature map $\mathbf V^*\in\mathbb R^{H_v\times W_v\times d_v}$, we apply $4\times4$ adaptive average pooling and flatten the pooled features in row-major order to obtain fixed-length reference features:
\begin{equation}
\mathbf F^*
=
\operatorname{Flatten}_{\mathrm{row}}
\left(
\operatorname{AdaptiveAvgPool}_{4\times4}(\mathbf V^*)
\right)
\in\mathbb R^{16\times d_v},
\end{equation}
where $N_p=16$ denotes the number of pooled spatial positions, and $d_v$ is the visual feature dimension.
For visual reconstruction, the auxiliary decoder uses an eight-head cross-attention layer with output dimension $d_v$, where fixed $4\times4$ two-dimensional sine-cosine positional embeddings $\mathbf P$ serve as queries and the corresponding region states $\mathbf Z_{t,n}$ serve as keys and values:
\begin{equation}
\hat{\mathbf F}_{t,n}
=
\operatorname{LN}
\left(
\mathbf P+
\operatorname{MHA}(\mathbf P,\mathbf Z_{t,n},\mathbf Z_{t,n})
\right).
\end{equation}
The visual supervision combines position-wise cosine distance and mean squared error:
\begin{equation}
\mathcal L_{\mathrm{visual}}^{t,n}
=
\frac{\lambda_{\mathrm{cos}}}{N_p}
\sum_{j=1}^{N_p}
\left[
1-\cos
\left(
\hat{\mathbf F}_{t,n,j},
\mathbf F^*_{t,n,j}
\right)
\right]
+
\frac{\lambda_{\mathrm{mse}}}{N_p d_v}
\left\|
\hat{\mathbf F}_{t,n}-\mathbf F^*_{t,n}
\right\|_F^2,
\end{equation}
where $\lambda_{\mathrm{cos}}=\lambda_{\mathrm{mse}}=1$.
For spatial supervision, the prediction head concatenates the $M$ latents of each region state and predicts normalized bounding-box coordinates through an MLP followed by a sigmoid function.
The horizontal and vertical coordinate pairs are sorted to ensure ordered box coordinates.
The spatial loss combines the $L_1$ loss over normalized coordinates with the GIoU loss:
\begin{equation}
\mathcal L_{\mathrm{spatial}}^{t,n}
=
\frac{1}{4}
\sum_{k=1}^{4}
\left|
\hat b_{t,n,k}-b^*_{t,n,k}
\right|
+
1-\operatorname{GIoU}
\left(
\hat{\mathbf b}_{t,n},
\mathbf b^*_{t,n}
\right).
\end{equation}
The auxiliary decoder is used only during training. At inference time, it is removed and TAPD decodes the final bounding boxes.

\noindent\textbf{Trajectory Generation.}
At each grounding step, the model first generates textual reasoning, followed by one or more region states.
Each region state corresponds to a single target and contains $M=2$ continuous region latents, delimited by the special tokens \texttt{<|latent\_start|>} and \texttt{<|latent\_end|>}.
Upon completing all region states for the current step, the model terminates the region sequence and subsequently resumes textual reasoning.
Consequently, the predicted number of targets $\hat{N}_t$ at step $t$ equals the number of region states generated at that step.
TAPD then instantiates a coordinate block for each region state and decodes the corresponding bounding box.
During training, targets within each grounding step are serialized according to their annotation order, whereas during inference, region states are parsed in generation order.
Evaluation is performed by matching predicted and ground-truth bounding box sets within each step without requiring order correspondence, thereby accounting for both missed and redundant predictions.

\begingroup
\newcommand{\tapdalgrow}[3]{#1: & \hspace*{#2em}\parbox[t]{\dimexpr\linewidth-#2em\relax}{\raggedright\strut #3\strut}\\}
\setlength{\parindent}{0pt}
\setlength{\intextsep}{1\baselineskip}

\begin{algorithm}[H]
\caption{Training of TAPD}
\label{alg:tapd_training}
\textbf{Input:} $p_\theta$; image $I$; task input $S$; trajectory $\mathcal H$; ground-truth boxes $\{\mathbf b_{t,n}^\ast\}$.
\par\smallskip

\begingroup
\renewcommand{\arraystretch}{1.06}
\setlength{\tabcolsep}{0pt}
\begin{tabularx}{\linewidth}{@{}r@{\hspace{0.7em}}X@{}}
\tapdalgrow{1}{0}{\textbf{for each} $(t,n)$ \textbf{do}}
\tapdalgrow{2}{1.3}{$\mathcal B_{t,n}\gets\operatorname{Encode}(\mathbf b_{t,n}^\ast)$ \quad\textit{// three digits per coordinate}}
\tapdalgrow{3}{1.3}{Sample masked token positions $\mathcal M_{t,n}$ by coordinate, leaving $0$--$3$ coordinates visible.}
\tapdalgrow{4}{1.3}{$\widetilde{\mathcal B}_{t,n}\gets\mathcal B_{t,n};\quad \widetilde{\mathcal B}_{t,n}[\mathcal M_{t,n}]\gets[\mathrm{MASK}]$}
\tapdalgrow{5}{0}{\textbf{end for}}
\tapdalgrow{6}{0}{Construct $\widetilde{\mathcal H}$ from $\mathcal H$ and $\{\widetilde{\mathcal B}_{t,n}\}$ (Eq.~(\ref{eq:expanded_trajectory})).}
\tapdalgrow{7}{0}{Estimate $\mathcal L_{\mathrm{box}}$ under the trajectory-aware attention mask (Eq.~(\ref{eq:loss_box})).}
\tapdalgrow{8}{0}{Optimize the joint SFT objective $\mathcal L_{\mathrm{SFT}}$ (Eq.~(\ref{eq:loss_sft})).}
\end{tabularx}
\endgroup

\end{algorithm}

\begin{algorithm}[H]
\caption{Inference of TAPD}
\label{alg:tapd_inference}
\textbf{Input:} $p_\theta$; image $I$; task input $S$; trajectory $\mathcal H$.
\par\smallskip

\begingroup
\renewcommand{\arraystretch}{1.06}
\setlength{\tabcolsep}{0pt}
\begin{tabularx}{\linewidth}{@{}r@{\hspace{0.7em}}X@{}}
\tapdalgrow{1}{0}{$\mathcal B_{t,n}\gets[\mathrm{MASK}]^{12},\quad\forall(t,n)$}
\tapdalgrow{2}{0}{\textbf{for} $r=1,\ldots,4$ \textbf{do}}
\tapdalgrow{3}{1.3}{Construct $\widetilde{\mathcal H}^{(r)}$ from $\mathcal H$ and the current $\{\mathcal B_{t,n}\}$ (Eq.~(\ref{eq:expanded_trajectory})).}
\tapdalgrow{4}{1.3}{Compute $\{p_{r,c,j}\}$ for all blocks in one forward pass using the trajectory-aware attention mask (Eq.~(\ref{eq:parallel_factorization})).}
\tapdalgrow{5}{1.3}{\textbf{for each} $\mathcal B_{t,n}$ \textbf{in parallel do}}
\tapdalgrow{6}{2.6}{$\mathcal C_r\gets\{c:\mathcal B_{t,n}[c]=[\mathrm{MASK}]^3\}$ \quad\textit{// $\mathcal B[c]$: three digit tokens}}
\tapdalgrow{7}{2.6}{Determine $\mathcal I_r(c)$ from recovered coordinates and box constraints, $c\in\mathcal C_r$.}
\tapdalgrow{8}{2.6}{$\mathcal V_r(c)\gets\mathcal I_r(c)\cap\{0,\ldots,999\},\quad c\in\mathcal C_r$}
\tapdalgrow{9}{2.6}{$S_r(c,v)\gets\sum_{j=1}^{3}\log p_{r,c,j}(d_j(v)),\quad c\in\mathcal C_r,\ v\in\mathcal V_r(c)$}
\tapdalgrow{10}{2.6}{$\widehat v_r(c)\gets\operatorname*{arg\,max}_{v\in\mathcal V_r(c)}S_r(c,v),\quad c\in\mathcal C_r$\quad\textit{// best value per coordinate}}
\tapdalgrow{11}{2.6}{$c_r^\ast\gets\operatorname*{arg\,max}_{c\in\mathcal C_r}S_r(c,\widehat v_r(c))$\quad\textit{// highest-confidence coordinate}}
\tapdalgrow{12}{2.6}{$\mathcal B_{t,n}[c_r^\ast]\gets\bigl(d_j(\widehat v_r(c_r^\ast))\bigr)_{j=1}^{3}$}
\tapdalgrow{13}{1.3}{\textbf{end for}}
\tapdalgrow{14}{0}{\textbf{end for}}
\end{tabularx}
\endgroup

\par\smallskip
\textbf{Output:} Boxes decoded from $\{\mathcal B_{t,n}\}$, grouped by step $t$.
\end{algorithm}

\endgroup

\noindent\textbf{Training and Inference of TAPD.}
During SFT, we randomly keep zero to three coordinates visible in each ground-truth block and mask the rest.
At inference, all blocks are decoded in parallel over four rounds, recovering the highest-confidence geometrically feasible coordinate per block at each round.
Algorithms~\ref{alg:tapd_training} and~\ref{alg:tapd_inference} summarize TAPD training and inference.

\noindent\textbf{GRPO Policy Update.}
Following the GRPO framework~\citep{shao2024deepseekmath}, for each input $q$, we sample $G$ trajectories $\{o_i\}_{i=1}^{G}$ from $\pi_{\theta_{\mathrm{old}}}$.
We apply GRPO to the sampled discrete-token actions within each trajectory, denoting their positions by $\mathcal T_i$.
For each $j\in\mathcal T_i$, the token-level probability ratio is defined as:
\begin{equation}
r_{i,j}(\theta)
=\frac{\pi_\theta(o_{i,j}\mid q,\mathcal H_{\theta,i,<j})}
{\pi_{\theta_{\mathrm{old}}}(o_{i,j}\mid q,\mathcal H_{\theta_{\mathrm{old}},i,<j})},
\label{eq:grpo_action_likelihood}
\end{equation}
where $\mathcal H_{\theta,i,<j}$ represents the text-latent history deterministically reconstructed from the same discrete prefix under the corresponding policy.

Given the overall reward, $R_i=R(o_i)$ defined in the main text, the group-relative advantage is computed as:
\begin{equation}
\hat A_i
=\frac{R_i-\operatorname{mean}(\{R_k\}_{k=1}^{G})}
{\operatorname{std}(\{R_k\}_{k=1}^{G})+\delta},
\label{eq:grpo_advantage}
\end{equation}
where $\delta>0$ is a numerical stability constant, and $\hat A_i$ is shared across all sampled tokens in $o_i$.
The policy is optimized using token-level clipping and per-trajectory averaging:
\begin{equation}
\begin{aligned}
\mathcal J_{\mathrm{GRPO}}(\theta)
={}&\mathbb E_{\substack{q\sim\mathcal D,\\
\{o_i\}_{i=1}^{G}\sim\pi_{\theta_{\mathrm{old}}}(\cdot\mid q)}}
\Bigg[\frac1G\sum_{i=1}^{G}\frac1{|\mathcal T_i|}\sum_{j\in\mathcal T_i}
\Big\{\\[-2pt]
&\min\!\left[
r_{i,j}(\theta)\hat A_i,\,
\operatorname{clip}\!\left(r_{i,j}(\theta),1-\varepsilon,1+\varepsilon\right)\hat A_i
\right]
-\beta D_{\mathrm{KL}}^{i,j}(\theta)
\Big\}\Bigg],
\end{aligned}
\label{eq:grpo_policy_objective}
\end{equation}
where $D_{\mathrm{KL}}^{i,j}$ denotes the token-level KL divergence relative to the fixed SFT reference policy.
TAPD decoding is employed solely to recover bounding boxes for reward computation, and 
the sampled rollouts from the fixed prefix are used  exclusively to estimate $R_{\mathrm{TR}}$, where the policy updates are applied to the original $G$ sampled trajectories.

\subsection{Evaluation on General REC Benchmarks}
\label{app:rec_evaluation}
For RefCOCO, RefCOCO+, RefCOCOg, Ref-L4, Ref-Adv, and GroundingME, we report Acc@0.5, which counts a predicted box as correct when its IoU with the ground-truth box is at least 0.5.
For HumanRef, we report Density-F1 (DF1), averaged over IoU thresholds from 0.5 to 0.95.
For PR-Bench, we report Mean Accuracy (mAcc), averaged over its five grounding categories and IoU thresholds from 0.5 to 0.9.
\section{In-Depth Analysis of ReferFlow-Bench}
\label{app:indepth_bench}

\subsection{Detailed Results across Eight Trajectory Subtypes}
\label{app:bench_subtypes}
Tab.~\ref{tab:bench_subtypes_refine_shift} and Tab.~\ref{tab:bench_subtypes_reset_mixture} present the results across the eight trajectory subtypes.
In the Refinement subtype, Attribute consistently outperforms Relational, suggesting that target-specific attributes provide more effective constraints for candidate filtering than relations involving other objects.
Similarly, Spatial generally outperforms Semantic in the Shift subtype, indicating that explicit spatial relations support transitions from previously grounded regions more effectively than semantic associations.
In the Reset subtype, Cross-category consistently achieves higher performance than Intra-category, even though both require independent localization.
This performance gap suggests that category changes help separate new targets from historical regions, whereas shared category semantics increase interference from previous grounding results.
In the Mixture subtype, Sequential consistently outperforms Interleaved, demonstrating that directly continuing the preceding grounding step is more reliable than selectively reusing earlier grounding results.
This highlights the importance of identifying relevant historical regions rather than merely maintaining the most recent region state.
ReferFlow achieves the best performance across all subtypes, demonstrating its ability to exploit relevant region information while limiting interference from irrelevant historical states in each step.

\begin{table}[p]
\centering
\caption{Performance comparison on the Refinement and Shift subtypes of ReferFlow-Bench.}
\label{tab:bench_subtypes_refine_shift}
\fontsize{8}{8.7}\selectfont
\setlength{\tabcolsep}{0.7pt}
\renewcommand{\arraystretch}{1.0}
\begin{tabular*}{\linewidth}{@{\extracolsep{\fill}}lc rrr rrr rrr rrr@{}}
\toprule
\multirow{3}{*}{\textbf{Model}} & \multirow{3}{*}{\textbf{Size}} & \multicolumn{6}{c}{\textbf{Refinement}} & \multicolumn{6}{c}{\textbf{Shift}} \\
\cmidrule(lr){3-8}\cmidrule(lr){9-14}
& & \multicolumn{3}{c}{\textbf{Attribute}} & \multicolumn{3}{c}{\textbf{Relational}} & \multicolumn{3}{c}{\textbf{Spatial}} & \multicolumn{3}{c}{\textbf{Semantic}} \\
\cmidrule(lr){3-5}\cmidrule(lr){6-8}\cmidrule(lr){9-11}\cmidrule(lr){12-14}
& & Q-Acc & T-Acc & F-Acc & Q-Acc & T-Acc & F-Acc & Q-Acc & T-Acc & F-Acc & Q-Acc & T-Acc & F-Acc \\
\midrule
Gemini 3.1 Pro & -- & 28.5 & 11.7 & 50.7 & 22.4 & 7.2 & 44.3 & 43.1 & 17.5 & 35.5 & 39.6 & 14.9 & 31.4 \\
GPT-5.6 Sol & -- & 36.0 & 12.8 & 53.4 & 30.9 & 9.1 & 47.9 & 52.8 & 24.7 & 39.0 & 49.9 & 22.7 & 35.5 \\
\midrule
Ovis2.5 & 2B & 21.1 & 3.4 & 30.0 & 15.4 & 0.3 & 23.8 & 37.8 & 8.0 & 24.5 & 34.6 & 5.7 & 20.7 \\
Qwen3-VL & 2B & 24.8 & 5.0 & 33.5 & 18.2 & 0.5 & 26.5 & 45.6 & 13.2 & 29.7 & 41.8 & 10.6 & 25.6 \\
Qwen3.5 & 4B & 29.6 & 5.1 & 37.4 & 24.7 & 1.6 & 32.1 & 48.3 & 15.6 & 35.9 & 45.7 & 13.6 & 33.0 \\
LLaVA-OneVision-1.5 & 4B & 12.5 & 0.7 & 10.6 & 6.4 & 0.1 & 4.0 & 28.0 & 7.2 & 21.4 & 24.5 & 4.9 & 17.3 \\
Youtu-VL & 4B & 12.9 & 2.6 & 23.5 & 7.8 & 0.3 & 17.8 & 46.0 & 10.0 & 26.8 & 43.1 & 7.9 & 23.3 \\
MiMo-VL-RL & 7B & 21.4 & 2.8 & 23.8 & 15.7 & 0.3 & 17.5 & 39.3 & 8.3 & 28.7 & 36.1 & 6.0 & 24.9 \\
Keye-VL-1.5 & 8B & 18.5 & 1.4 & 16.8 & 11.9 & 0.2 & 9.8 & 38.5 & 5.4 & 19.4 & 34.7 & 2.8 & 15.3 \\
InternVL3.5 & 8B & 21.0 & 6.0 & 24.4 & 16.1 & 2.5 & 19.1 & 40.0 & 10.3 & 27.6 & 37.4 & 8.3 & 24.7 \\
Qwen3.5 & 9B & 31.3 & 5.4 & 39.3 & 25.2 & 0.9 & 32.7 & 52.6 & 16.6 & 38.3 & 49.1 & 14.3 & 34.2 \\
GLM-4.6V-Flash & 9B & 28.8 & 7.3 & 44.8 & 23.7 & 3.6 & 39.3 & 53.7 & 20.7 & 38.0 & 50.8 & 18.6 & 34.5 \\
Qwen3.5 & 27B & 39.1 & 7.0 & 53.7 & 33.4 & 2.7 & 47.6 & 57.8 & 19.5 & 41.9 & 54.6 & 17.2 & 38.1 \\
Qwen3.8 & 27B & 40.5 & 7.1 & 55.1 & 33.9 & 2.4 & 48.1 & 55.6 & 19.3 & 39.5 & 51.8 & 16.7 & 35.4 \\
Ovis2.6 & 30B-A3B & 31.2 & 10.4 & 46.1 & 26.3 & 6.9 & 40.8 & 54.2 & 16.7 & 39.4 & 51.6 & 14.7 & 36.5 \\
InternVL3.5 & 38B & 27.6 & 10.6 & 42.0 & 21.5 & 6.1 & 35.5 & 51.2 & 23.6 & 39.8 & 47.7 & 21.3 & 35.7 \\
GLM-4.6V & 106B-A12B & 34.9 & 7.9 & 43.9 & 29.8 & 4.2 & 38.4 & 56.6 & 17.9 & 39.2 & 53.7 & 15.8 & 35.7 \\
Qwen3.5 & 122B-A10B & 40.3 & 9.4 & 55.4 & 34.6 & 5.1 & 49.3 & 57.6 & 20.4 & 41.5 & 54.4 & 18.1 & 37.7 \\
Step-3.7-Flash & 198B-A11B & 31.9 & 12.6 & 39.5 & 25.2 & 7.9 & 32.5 & 54.6 & 23.2 & 46.5 & 50.8 & 20.6 & 42.4 \\
InternVL3.5 & 241B-A28B & 26.0 & 8.6 & 37.7 & 21.1 & 5.1 & 32.4 & 53.4 & 21.5 & 41.7 & 50.8 & 19.5 & 38.8 \\
\midrule
DeepEyes & 7B & 15.4 & 4.5 & 23.6 & 9.3 & 0.4 & 17.0 & 33.8 & 5.4 & 15.6 & 30.3 & 3.1 & 11.5 \\
RegionReasoner & 7B & 17.5 & 0.0 & 10.6 & 12.4 & 0.0 & 5.1 & 37.8 & 9.3 & 24.8 & 34.9 & 7.3 & 21.3 \\
Rex-Omni & 3B & 46.7 & 16.4 & 53.7 & 41.0 & 12.1 & 47.6 & 59.2 & 21.4 & 44.1 & 56.0 & 19.1 & 40.3 \\
\midrule
\textbf{ReferFlow} & \textbf{2B} & \textbf{59.1} & \textbf{24.7} & \textbf{68.8} & \textbf{53.6} & \textbf{20.6} & \textbf{62.9} & \textbf{70.8} & \textbf{36.8} & \textbf{63.4} & \textbf{67.9} & \textbf{34.5} & \textbf{59.9} \\
\bottomrule
\end{tabular*}
\end{table}

\begin{table}[p]
\centering
\caption{Performance comparison on the Reset and Mixture subtypes of ReferFlow-Bench.}
\label{tab:bench_subtypes_reset_mixture}
\fontsize{8}{8.7}\selectfont
\setlength{\tabcolsep}{0.7pt}
\renewcommand{\arraystretch}{1.0}
\begin{tabular*}{\linewidth}{@{\extracolsep{\fill}}lc rrr rrr rrr rrr@{}}
\toprule
\multirow{3}{*}{\textbf{Model}} & \multirow{3}{*}{\textbf{Size}} & \multicolumn{6}{c}{\textbf{Reset}} & \multicolumn{6}{c}{\textbf{Mixture}} \\
\cmidrule(lr){3-8}\cmidrule(lr){9-14}
& & \multicolumn{3}{c}{\textbf{Cross-category}} & \multicolumn{3}{c}{\textbf{Intra-category}} & \multicolumn{3}{c}{\textbf{Sequential}} & \multicolumn{3}{c}{\textbf{Interleaved}} \\
\cmidrule(lr){3-5}\cmidrule(lr){6-8}\cmidrule(lr){9-11}\cmidrule(lr){12-14}
& & Q-Acc & T-Acc & F-Acc & Q-Acc & T-Acc & F-Acc & Q-Acc & T-Acc & F-Acc & Q-Acc & T-Acc & F-Acc \\
\midrule
Gemini 3.1 Pro & -- & 36.2 & 6.6 & 24.9 & 30.9 & 3.2 & 19.6 & 34.4 & 6.6 & 34.4 & 27.3 & 1.5 & 27.0 \\
GPT-5.6 Sol & -- & 48.5 & 9.4 & 34.2 & 44.3 & 6.6 & 29.7 & 50.2 & 9.3 & 47.9 & 44.2 & 5.1 & 41.6 \\
\midrule
Ovis2.5 & 2B & 31.4 & 1.8 & 29.7 & 26.3 & 0.1 & 24.6 & 28.9 & 4.0 & 26.8 & 22.2 & 0.4 & 19.7 \\
Qwen3-VL & 2B & 41.8 & 7.1 & 32.9 & 36.5 & 3.5 & 27.3 & 40.2 & 6.7 & 32.8 & 32.8 & 1.3 & 24.8 \\
Qwen3.5 & 4B & 47.6 & 9.1 & 31.1 & 43.7 & 6.6 & 26.9 & 38.1 & 10.1 & 33.3 & 32.5 & 6.1 & 27.3 \\
LLaVA-OneVision-1.5 & 4B & 21.3 & 0.9 & 2.6 & 16.0 & 0.1 & 0.1 & 21.3 & 0.0 & 4.3 & 14.2 & 0.0 & 0.1 \\
Youtu-VL & 4B & 39.2 & 5.2 & 21.0 & 35.0 & 2.4 & 16.5 & 32.4 & 3.4 & 26.2 & 26.4 & 0.3 & 19.9 \\
MiMo-VL-RL & 7B & 32.7 & 3.7 & 22.5 & 27.6 & 0.3 & 17.4 & 32.2 & 2.5 & 22.3 & 25.5 & 0.3 & 15.2 \\
Keye-VL-1.5 & 8B & 33.2 & 2.2 & 14.7 & 27.9 & 0.2 & 9.1 & 32.0 & 1.5 & 21.5 & 24.6 & 0.2 & 13.5 \\
InternVL3.5 & 8B & 30.7 & 3.1 & 22.6 & 26.8 & 0.6 & 18.4 & 34.0 & 6.3 & 27.3 & 28.4 & 2.3 & 21.3 \\
Qwen3.5 & 9B & 50.4 & 10.1 & 33.7 & 45.1 & 6.7 & 28.4 & 38.9 & 12.0 & 41.5 & 31.8 & 6.9 & 34.1 \\
GLM-4.6V-Flash & 9B & 48.4 & 9.4 & 32.9 & 44.2 & 6.6 & 28.4 & 47.7 & 12.1 & 45.5 & 41.7 & 7.9 & 39.2 \\
Qwen3.5 & 27B & 53.5 & 10.9 & 38.9 & 48.4 & 7.5 & 33.9 & 49.6 & 18.7 & 52.7 & 42.9 & 14.0 & 45.6 \\
Qwen3.8 & 27B & 54.5 & 11.8 & 40.0 & 49.2 & 8.1 & 34.4 & 51.8 & 14.4 & 52.4 & 44.4 & 9.3 & 44.4 \\
Ovis2.6 & 30B-A3B & 45.5 & 8.0 & 31.3 & 41.6 & 5.5 & 27.1 & 45.1 & 9.2 & 43.5 & 39.5 & 5.2 & 37.5 \\
InternVL3.5 & 38B & 44.7 & 9.4 & 28.7 & 39.4 & 6.0 & 23.4 & 46.5 & 10.4 & 51.7 & 39.4 & 5.3 & 44.3 \\
GLM-4.6V & 106B-A12B & 51.6 & 10.5 & 34.3 & 47.4 & 7.7 & 29.8 & 50.3 & 7.4 & 48.8 & 44.3 & 3.2 & 42.5 \\
Qwen3.5 & 122B-A10B & 52.3 & 11.4 & 35.7 & 47.2 & 8.0 & 30.6 & 46.8 & 8.5 & 48.2 & 40.1 & 3.8 & 41.1 \\
Step-3.7-Flash & 198B-A11B & 47.8 & 10.7 & 34.0 & 42.5 & 7.0 & 28.4 & 37.8 & 7.3 & 40.5 & 30.4 & 2.2 & 32.5 \\
InternVL3.5 & 241B-A28B & 50.7 & 11.2 & 35.8 & 46.8 & 8.7 & 31.6 & 48.0 & 10.4 & 46.8 & 42.4 & 6.4 & 40.8 \\
\midrule
DeepEyes & 7B & 29.2 & 8.6 & 21.1 & 23.9 & 5.2 & 15.8 & 32.6 & 2.8 & 23.6 & 25.5 & 0.6 & 16.2 \\
RegionReasoner & 7B & 39.9 & 5.9 & 24.4 & 35.7 & 3.1 & 20.0 & 27.9 & 4.7 & 25.6 & 21.9 & 0.5 & 19.4 \\
Rex-Omni & 3B & 50.4 & 13.3 & 38.6 & 45.3 & 10.0 & 33.6 & 57.3 & 8.7 & 37.2 & 50.6 & 4.0 & 30.1 \\
\midrule
\textbf{ReferFlow} & \textbf{2B} & \textbf{67.7} & \textbf{30.8} & \textbf{65.1} & \textbf{63.2} & \textbf{27.7} & \textbf{60.3} & \textbf{66.2} & \textbf{27.2} & \textbf{63.4} & \textbf{59.9} & \textbf{22.7} & \textbf{56.7} \\
\bottomrule
\end{tabular*}
\end{table}

\subsection{Single-Turn vs. Multi-Turn Evaluation}
\label{app:bench_protocol}

\begin{figure}[htbp]
    \centering
    \includegraphics[width=\linewidth]{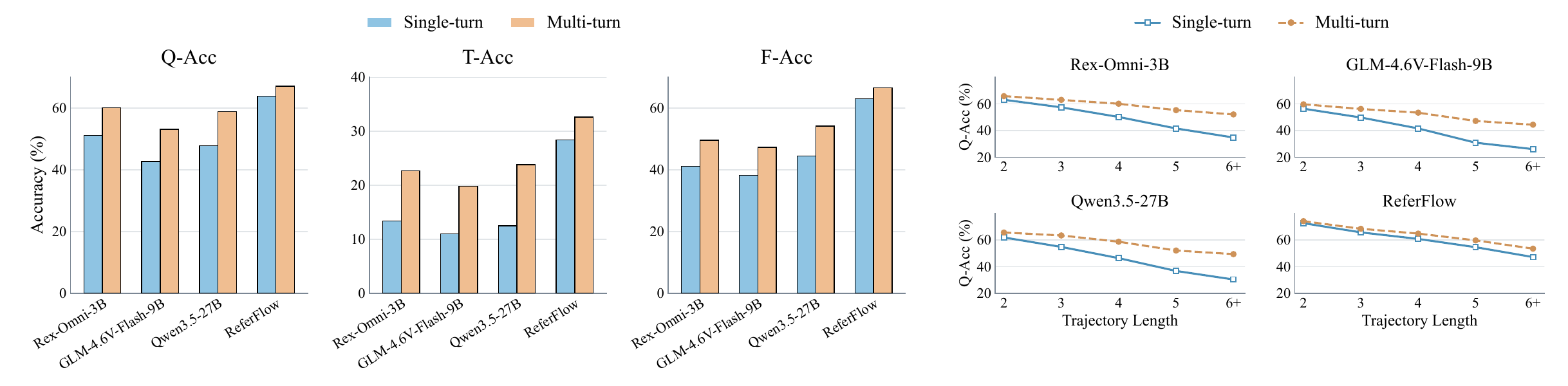}
    \caption{Comparisons between multi-turn and single-turn evaluation protocols on ReferFlow-Bench.}
    \label{fig:bench_protocol}
\end{figure}

In our default single-turn setting, the model receives the complete ordered sequence of grounding steps at once and predicts their target regions in one response. We compare this with a multi-turn setting that presents one step-level query per turn. This setting retains the original image and full dialogue history, including the model’s own predicted bounding boxes.
The single-turn setting requires models to track grounding steps and their corresponding target regions throughout generation, whereas the multi-turn setting explicitly specifies the current query at each turn.
As shown in Fig.~\ref{fig:bench_protocol}, the single-turn setting yields lower Q-Acc and T-Acc than the multi-turn setting. Moreover, Q-Acc declines more sharply with trajectory length in the single-turn setting, widening the gap between the two settings. Multi-turn evaluation explicitly specifies the current query at each turn, guiding the model through successive grounding steps. In contrast, single-turn evaluation requires the model to track its progress through the complete query sequence while grounding the targets at each step. These results suggest that single-turn evaluation better reveals the challenges of consistently grounding intermediate visual references within a complex query without turn-by-turn guidance.

\subsection{Effects of Intermediate Grounding}
\label{app:intermediate_grounding}
\begin{figure}[!htbp]
    \centering
    \includegraphics[width=\linewidth]{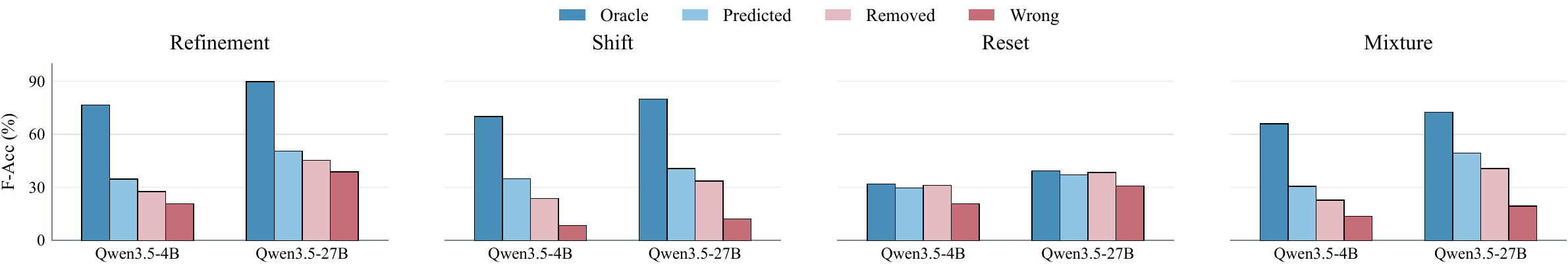}
    \caption{Effects of intermediate grounding on final localization across trajectory types.}
    \label{fig:intermediate_box_intervention}
\end{figure}

We investigate the impact of intermediate grounding on final localization across the full ReferFlow-Bench, reporting F-Acc in Fig.~\ref{fig:intermediate_box_intervention}.
Prior to predicting the final target, we either retain the predicted intermediate bounding boxes (Predicted), replace them with ground-truth bounding boxes (Oracle) or distractor bounding boxes (Wrong), or omit them entirely (Removed).
The input image and the entire query sequence remain fixed across all conditions.
Across both models, the Oracle condition consistently outperforms the Predicted condition for the Refinement, Shift, and Mixture trajectory types.
Furthermore, the Wrong condition underperforms the Removed condition across all four trajectory types, suggesting that incorrect intermediate bounding boxes can negatively interfere with final localization.
Shift shows the largest gap between Oracle and Wrong in both models, while Reset remains relatively stable, reflecting their different dependencies on prior regions.

\section{In-Depth Analysis of ReferFlow}
\label{app:indepth_model}

\subsection{Efficiency Analysis}
\label{app:model_efficiency}

Tab.~\ref{tab:app_model_efficiency} compares ReferFlow against representative models across three benchmarks, evaluating the average wall-clock time, output sequence length, and grounding accuracy. 
We randomly sample 800 instances each from RefCOCOg and GroundingME, and 800 trajectories from ReferFlow-Bench, each containing five ground-truth boxes across all steps. 
LocateAnything is excluded from ReferFlow-Bench because its current implementation does not support grounding an entire trajectory within a single input. 
The longer output sequences on GroundingME partly reflect its longer queries, as the query text is repeated in the generated responses.
Compared to the same-backbone baseline Qwen3-VL-2B, ReferFlow improves grounding accuracy while reducing inference time and output sequence length across all benchmarks. 
SRLR utilizes compact continuous states to represent intermediate regions, while TAPD recovers all box coordinates in parallel. 
Overall, these components support trajectory-level grounding with competitive overall inference costs. 
ReferFlow also achieves highly competitive latency compared to LocateAnything, despite producing longer output sequences. 
Given that the two models differ in backbone and model size, these measurements compare overall system performance rather than isolated decoding efficiency.

To examine their decoding costs, we compare the sequential decoding steps required to generate $N$ boxes. 
This analysis accounts for region-state generation and box decoding steps, while excluding textual reasoning.

\textbf{(i) Autoregressive Token Decoding.}
Qwen3-VL represents coordinates as text and generates them autoregressively.
For coordinates in $0$--$999$, assuming one token per digit, each coordinate requires one to three predictions.
With five formatting tokens per box, generating $N$ boxes requires $9N$--$17N$ sequential predictions.

\textbf{(ii) Parallel Box Decoding with Fallback.} 
LocateAnything jointly predicts a six-token block containing four coordinates and two boundary tokens. 
Each block requires one parallel step, while decoding remains sequential across boxes. 
Accounting for six additional autoregressive predictions per fallback yields an expected box-decoding depth of $(1+6p)N$, where $p$ is the per-box fallback rate. 
In our evaluation, 24.5\% of RefCOCOg instances and 66.5\% of GroundingME instances triggered NTP fallback, indicating a greater reliance on autoregressive correction on the more challenging benchmark. 
Using these sample-level rates as per-box estimates yields approximately $2.47N$ and $4.99N$ decoding steps, respectively.

\textbf{(iii) ReferFlow.} 
ReferFlow decouples region-state generation from coordinate decoding. 
For each box, it generates two continuous latents delimited by \texttt{<|latent\_start|>} and \texttt{<|latent\_end|>}. 
TAPD then recovers all boxes in parallel over four geometry-constrained rounds. 
Accounting for four sequential steps per region state and four TAPD rounds yields a total of $4N+4$ steps. 
Thus, region-state generation scales with the box count, while coordinate recovery remains fixed at four rounds without NTP fallback.

Although ReferFlow achieves higher grounding accuracy, it requires more sequential decoding steps when predicting a small number of boxes according to these estimates. 
However, as the number of boxes increases, TAPD maintains a fixed coordinate-decoding depth of four rounds, enforcing geometric constraints without relying on cross-box autoregression or NTP fallback. 
These characteristics facilitate long grounding trajectories and highlight the potential of ReferFlow for other tasks that require predicting multiple bounding boxes, such as dense object detection.

\begin{table}[!t]
\centering
\caption{Efficiency analysis of ReferFlow compared with representative models. We report average wall-clock time (Time), sequence length (Len.), and accuracy (Acc.) across three benchmarks. For ReferFlow-Bench, Acc. denotes Q-Acc. Fallback denotes re-decoding unreliable parallel outputs using next-token prediction. -- denotes unsupported evaluation settings.}
\label{tab:app_model_efficiency}

\fontsize{8}{8.7}\selectfont
\setlength{\tabcolsep}{2.5pt}
\renewcommand{\arraystretch}{1.0}

\begin{tabular*}{\linewidth}{@{\extracolsep{\fill}}l ccc ccc ccc@{}}
\toprule
\multirow{2}{*}{\textbf{Model}}
& \multicolumn{3}{c}{\textbf{RefCOCOg}}
& \multicolumn{3}{c}{\textbf{GroundingME}}
& \multicolumn{3}{c}{\textbf{ReferFlow-Bench}} \\
\cmidrule(lr){2-4}\cmidrule(lr){5-7}\cmidrule(lr){8-10}
& Time (s) & Len. & Acc.
& Time (s) & Len. & Acc.
& Time (s) & Len. & Acc. \\
\midrule

Qwen3-VL-2B
& 1.5 & 29.4 & 84.3
& 2.8 & 85.6 & 27.6
& 6.3 & 237.1 & 34.8 \\

LocateAnything
& 0.9\,{\fontsize{6.5}{7.2}\selectfont
  \textcolor{black!65}{(24.5\% fallback)}}
& 10.2 & 88.0
& 1.9\,{\fontsize{6.5}{7.2}\selectfont
  \textcolor{black!65}{(66.5\% fallback)}}
& 53.7 & 23.9
& -- & -- & -- \\

\midrule
\textbf{ReferFlow}
& \textbf{0.8} & \textbf{18.8} & \textbf{92.2}
& \textbf{1.2} & \textbf{62.4} & \textbf{43.4}
& \textbf{1.8} & \textbf{119.4} & \textbf{63.2} \\
\bottomrule
\end{tabular*}
\end{table}

\label{app:bench_subtypes}

\subsection{Dependence on Historical Region States}
\begin{wrapfigure}{r}{0.35\textwidth}
\vspace{-6pt}
\centering
\includegraphics[width=\linewidth]{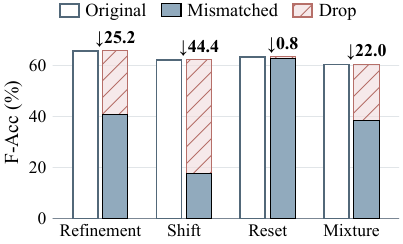}
\captionsetup{font=small,skip=3pt}
\caption{Effect of historical region states on final localization.}
\label{fig:historical_region_states}
\end{wrapfigure}

We replace the region state of one intermediate target with that of another object in the same image (Mismatched) and rerun the remaining grounding steps.
As shown in Fig.~\ref{fig:historical_region_states}, F-Acc drops substantially on Refinement, Shift, and Mixture, with the largest decrease on Shift.
In Shift, the target-specific visual and spatial information retained in a mismatched state can provide a misleading reference for locating the next target.
In contrast, Reset remains relatively stable, with an absolute F-Acc decrease of only 0.8\%.
For Reset, the small performance change suggests that ReferFlow limits interference from historical region states when localizing independently specified targets.
These results provide causal evidence that historical region states guide subsequent grounding, and suggest that ReferFlow selectively uses this information according to the task's dependence on prior regions.

\begin{figure}[!b]
    \centering
    \includegraphics[width=\linewidth]{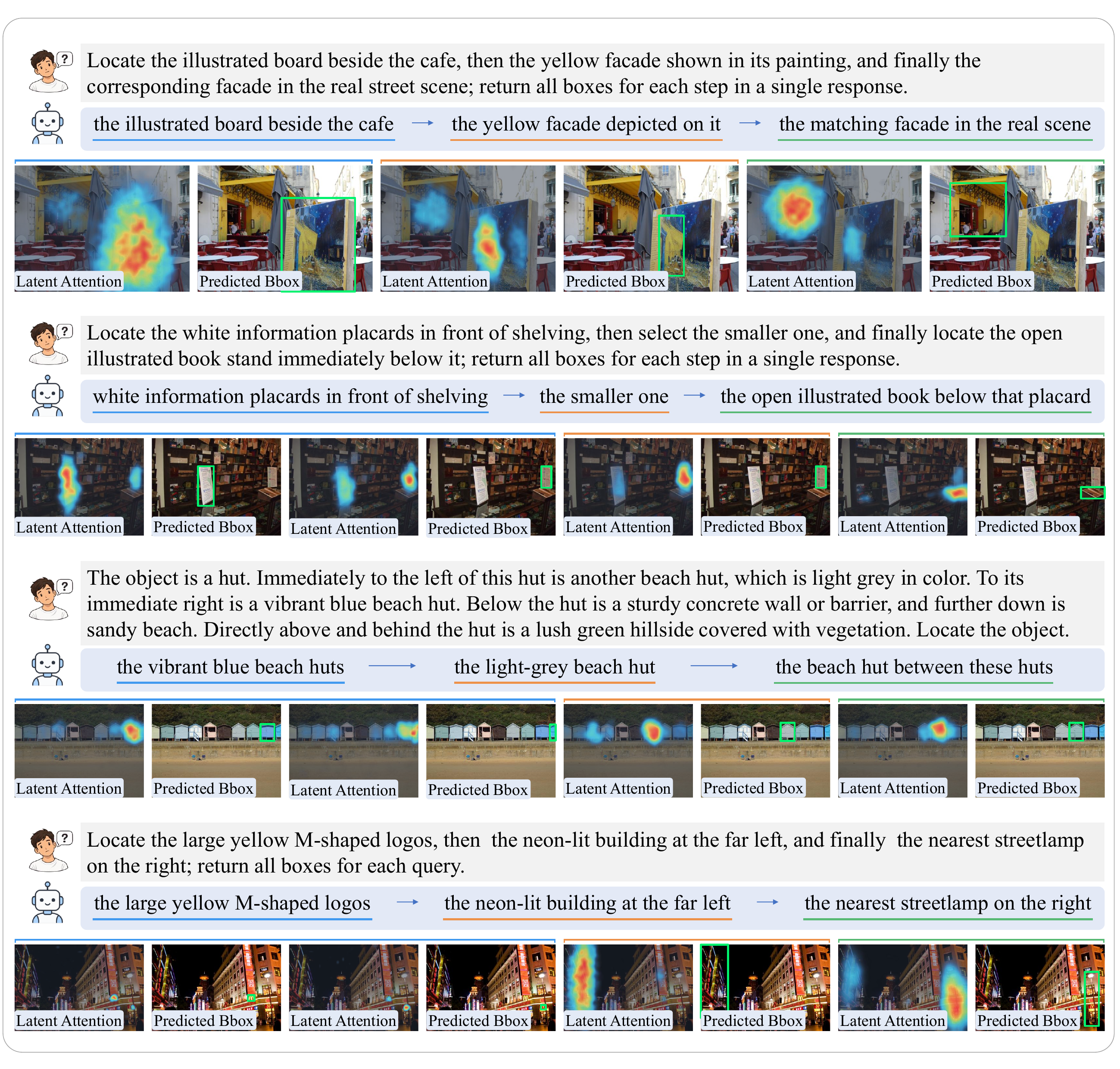}
    \caption{\textbf{Attention visualization of region latents.} Heatmaps depict the Gaussian-smoothed aggregation of relevant image patches for 
    region latents.}
    \label{fig:region_latent_attention}
\end{figure}
\subsection{Attention Visualization of Region States}
We visualize attention from region latents to image tokens alongside the corresponding textual responses and predicted boxes in Fig.~\ref{fig:region_latent_attention}.
For each region state, we average attention over its $M=2$ region latents and all heads in the last full-attention layer.
Across grounding steps, attention remains aligned with the corresponding target regions as constraints or targets change, providing qualitative evidence that region states capture target-specific visual information.

\subsection{Ablation on the Number of Region Latents}
\label{app:latent_length}
% Natural-height columns keep the table compact beside its discussion.
\noindent\begin{minipage}[t]{0.48\linewidth}
\vspace{0pt}
Table~\ref{tab:app_model_latent_length} evaluates the impact of the number of latents per region state.
Increasing $M$ from 1 to 2 improves performance across all evaluated benchmarks.
However, performance declines slightly at $M=4$ and drops further at $M=8$. This indicates that distributing the same region-level supervision across a larger number of latents may hinder region-state learning.
Therefore, we adopt $M=2$ as the default setting.
\end{minipage}\hfill
\begin{minipage}[t]{0.49\linewidth}
\vspace{-1pt}
\centering
\captionsetup{type=table,font=small,skip=3pt}
\caption{\textbf{Ablation on the number of region latents.} All variants are evaluated after SFT.}
\label{tab:app_model_latent_length}
\fontsize{8}{9.5}\selectfont
\setlength{\tabcolsep}{1pt}
\renewcommand{\arraystretch}{1.0}
\begin{tabular*}{\linewidth}{@{\extracolsep{\fill}}cccccc@{}}
\toprule
\multirow{2}{*}{$M$} & \multirow{2}{*}{\textbf{RefCOCOg}} & \multirow{2}{*}{\textbf{GroundingME}} & \multicolumn{3}{c}{\textbf{ReferFlow-Bench}} \\
\cmidrule(lr){4-6}
& & & Q-Acc & T-Acc & F-Acc \\
\midrule
1 & 91.3 & 37.6 & 59.0 & 23.7 & 57.5 \\
2 & 92.0 & 39.1 & 60.8 & 25.6 & 59.4 \\
4 & 91.9 & 38.7 & 60.4 & 25.0 & 58.6 \\
8 & 89.5 & 33.4 & 53.8 & 17.8 & 50.8 \\
\bottomrule
\end{tabular*}

\end{minipage}
\par

\subsection{MLP Regression vs. MLLM Coordinate Decoding}
% Keep the decoder plot beside the corresponding discussion.
\noindent\begin{minipage}[t]{0.64\linewidth}
\vspace{0pt}
We further compare the auxiliary MLP for SRLR training and TAPD at different IoU thresholds in Fig.~\ref{fig:decoder_comparison}.
At an IoU threshold of 0.3, the auxiliary MLP achieves comparable accuracy to TAPD on both benchmarks, indicating that region states retain spatial information for direct box prediction.
The performance gap widens at higher IoU thresholds.
Compared with the external MLP decoder, TAPD decodes coordinates within the MLLM, allowing it to combine the current region state with textual and regional evidence from the corresponding trajectory context for more precise localization.
\end{minipage}\hfill
\begin{minipage}[t]{0.33\linewidth}
\vspace{0pt}
\centering
\includegraphics[width=\linewidth]{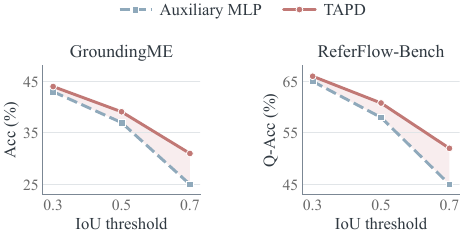}
\captionsetup{type=figure,font=small,skip=3pt,justification=raggedright,singlelinecheck=false}
\caption{Comparisons of coordinate decoding at different IoU thresholds after SFT.}
\label{fig:decoder_comparison}
\end{minipage}
\par

\subsection{Adaptive Trajectory Generation}
\label{app:adaptive_modes}
\begin{figure}
    \centering
    \includegraphics[width=0.6\linewidth]{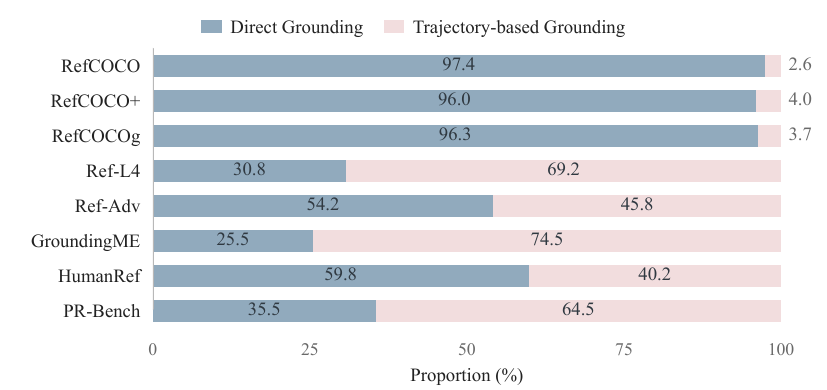}
    \caption{Proportions of direct and trajectory-based grounding across benchmarks.}
    \label{fig:app_model_adaptive_modes}
\end{figure}

 Fig.~\ref{fig:app_model_adaptive_modes} illustrates the proportion of trajectory generation by ReferFlow across various REC benchmarks. 
The results indicate that the utilization of grounding trajectories is closely correlated with task complexity.
On standard REC benchmarks such as RefCOCO/+/g, which are characterized by short queries with limited intermediate references, the model relies predominantly on direct grounding. 
In contrast, on Ref-L4, which focuses on long-caption grounding, trajectory-based grounding is employed more frequently. 
This pattern is most notable on GroundingME, where trajectory-based grounding accounts for 74.5\% of the predictions, reflecting a greater reliance on these trajectories for complex referring expressions. 
Additionally, Ref-Adv and HumanRef show more balanced distributions, though direct grounding remains more common.

\subsection{Effects of $R_{\mathrm{TR}}$}
We evaluate ReferFlow on the GroundingME benchmark following both SFT and RL, with and without the trajectory reward $R_{\mathrm{TR}}$.
To isolate the effect of $R_{\mathrm{TR}}$, $R_{\mathrm{multi}}$ is disabled in both RL settings.
We report the final localization accuracy, the average number of grounding steps, and the trajectory ratio, defined as the proportion of multi-step grounding outputs. 
As shown in Tab.~\ref{tab:app_trajectory_reward}(a), compared to SFT, RL without $R_{\mathrm{TR}}$ improves final localization accuracy but increases both the trajectory ratio and the average number of steps. 
Incorporating $R_{\mathrm{TR}}$ further increases the trajectory ratio and final localization accuracy while maintaining the average number of steps largely unchanged, demonstrating a more favorable trade-off between grounding accuracy and trajectory length. 
Furthermore, we manually evaluate intermediate grounding on 500 GroundingME samples, defining intermediate grounding accuracy as the proportion of intermediate steps that correctly ground their generated descriptions.
Tab.~\ref{tab:app_trajectory_reward}(b) reveals that $R_{\mathrm{TR}}$ enhances the accuracy of self-generated intermediate grounding steps, with performance gains observed across all trajectory length groups.

\begin{table}[htbp]
\centering
\captionsetup{font=footnotesize}
\caption{\textbf{Effects of trajectory reward on GroundingME.}
(a) Final localization accuracy, trajectory ratio, and average grounding steps, where trajectory ratio denotes the proportion of multi-step grounding outputs.
(b) Intermediate grounding accuracy overall and across different trajectory lengths.
}
\label{tab:app_trajectory_reward}
\scriptsize
\setlength{\tabcolsep}{3pt}
\renewcommand{\arraystretch}{1.15}
\begin{minipage}[t]{0.49\linewidth}
\centering
\textbf{(a) Final Accuracy and Trajectory Generation}\par\vspace{4pt}
\begin{tabular*}{\linewidth}{@{\extracolsep{\fill}}lccc@{}}
\toprule
\textbf{Setting} & \makecell{\textbf{Final}\\\textbf{Acc.(\%)}} & \makecell{\textbf{Trajectory}\\\textbf{Ratio (\%)}} & \makecell{\textbf{Avg.}\\\textbf{Steps}} \\
\midrule
SFT & 39.1 & 48.0 & 1.72 \\
RL w/o $R_{\mathrm{TR}}$ & 40.6 & 64.0 & 2.06 \\
RL w/ $R_{\mathrm{TR}}$ & \textbf{42.1} & 71.0 & 2.03 \\
\bottomrule
\end{tabular*}
\end{minipage}\hfill
\begin{minipage}[t]{0.49\linewidth}
\centering
\textbf{(b) Intermediate Accuracy by Trajectory Length}\par\vspace{4pt}
\begin{tabular*}{\linewidth}{@{\extracolsep{\fill}}lcccc@{}}
\toprule
\textbf{Setting} & \makecell{\textbf{Overall}\\\textbf{(\%)}} & \makecell{\textbf{2 steps}\\\textbf{(\%)}} & \makecell{\textbf{3 steps}\\\textbf{(\%)}} & \makecell{$\boldsymbol{\geq 4}$ \textbf{steps}\\\textbf{(\%)}} \\
\midrule
SFT & 64.2 & 66.2 & 63.0 & 56.0 \\
RL w/o $R_{\mathrm{TR}}$ & 65.6 & 68.1 & 65.0 & 58.7 \\
RL w/ $R_{\mathrm{TR}}$ & \textbf{71.8} & \textbf{73.6} & \textbf{70.4} & \textbf{64.0} \\
\bottomrule
\end{tabular*}
\end{minipage}
\end{table}

\subsection{Generalization to Object Detection}
\noindent\begin{minipage}[t]{0.58\linewidth}
\vspace{0pt}
To demonstrate its broader applicability, we further extend ReferFlow to object detection via supervised fine-tuning on the COCO and LVIS training sets, using vanilla SFT with text-based bounding-box outputs as a baseline. Tab.~\ref{tab:object_detection} shows that ReferFlow outperforms both the original Qwen3-VL-2B backbone and the vanilla SFT baseline in detection accuracy, while also achieving a lower average inference time. These results demonstrate ReferFlow's potential as an accurate and efficient framework for multi-object detection, extending its utility well beyond referring expression comprehension.
\end{minipage}\hfill
\begin{minipage}[t]{0.39\linewidth}
\vspace{0pt}
\begingroup
    \centering
    \captionsetup{type=table,font=small,skip=4pt}
    \caption{Object detection accuracy and inference efficiency on COCO and LVIS.}
    \label{tab:object_detection}
    \normalsize
    \setlength{\tabcolsep}{2pt}
    \renewcommand{\arraystretch}{1.85}
    \resizebox{\linewidth}{!}{%
    \begin{tabular}{@{}lccc@{}}
        \toprule
        \textbf{Method} & \makecell{\textbf{COCO}\\\textbf{(F1@mIoU)}} & \makecell{\textbf{LVIS}\\\textbf{(F1@mIoU)}} & \textbf{Time(s)} \\
        \midrule
        Qwen3-VL-2B
        & 46.2 & 40.7 & 2.5 \\
        \quad + Vanilla SFT
        & 52.8 & 43.0 & 2.2 \\
        \textbf{ReferFlow}
        & \textbf{55.1} & \textbf{43.8} & \textbf{1.1} \\
        \bottomrule
    \end{tabular}
    }
\par\endgroup

\end{minipage}
\par

% \section{Additional Experiments}\label{app:additional_exp}

% \subsection{Generalization across Backbones}
% Table~\ref{tab:backbone_generalization} reports the results of ReferFlow on additional backbones, including Qwen2.5-VL-3B and Qwen3-VL-4B.
% On both backbones, ReferFlow consistently outperforms the corresponding Text+NTP baselines across REC benchmarks and ReferFlow-Bench, following the improvement trend observed with Qwen3-VL-2B in the main paper and supporting its effectiveness across backbones and model scales.

\section{Visualization}
\label{app:visualization}

\subsection{Visualizations of ReferFlow-Bench}
\begin{figure}[htbp]
    \centering
    \includegraphics[width=\linewidth]{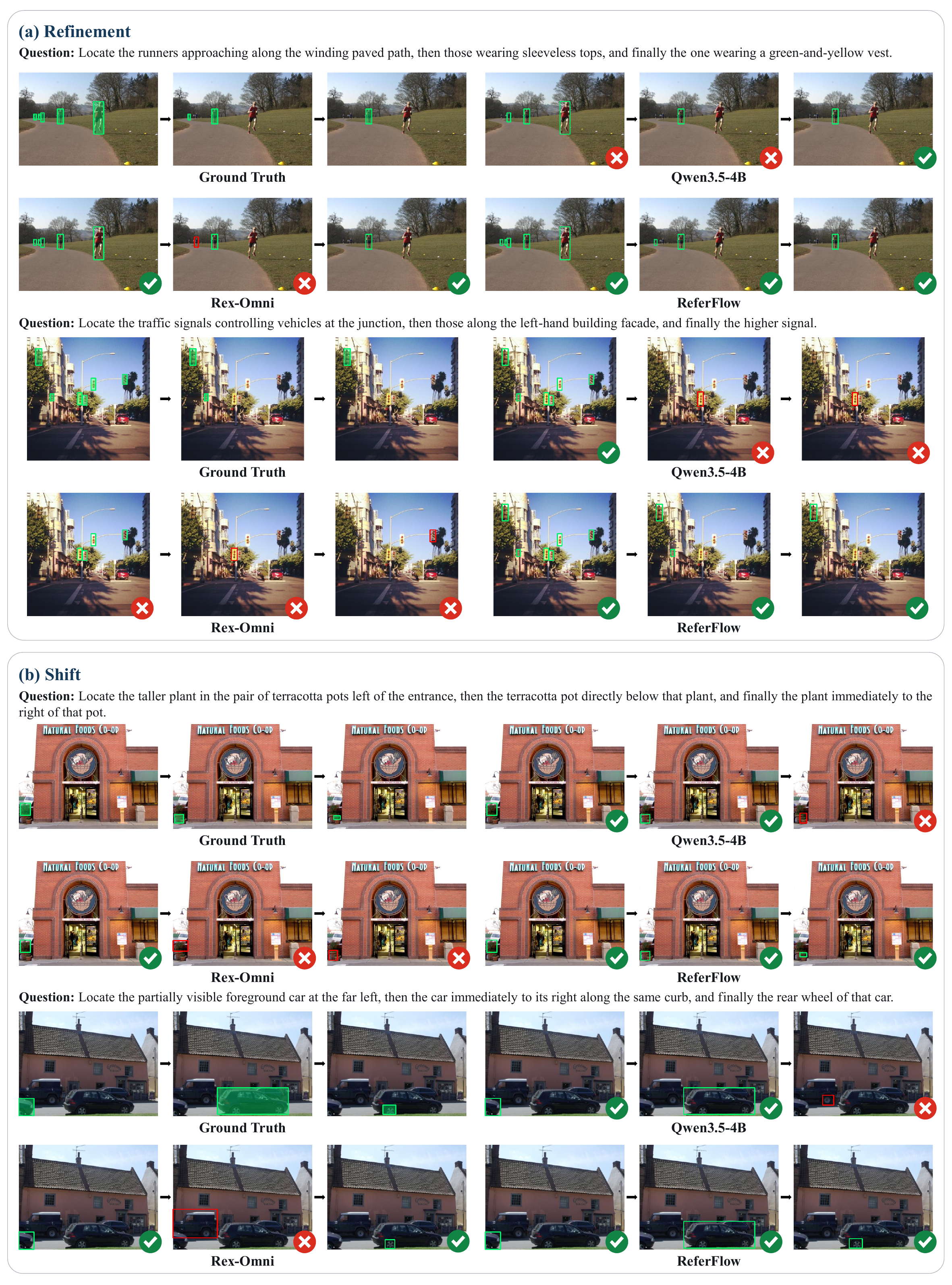}
    \caption{\textbf{Visualization results for Refinement and Shift on ReferFlow-Bench.}
Each panel presents a challenging example and the corresponding predictions from representative models.}
    \label{fig:bench_visualizations_refineshift}
\end{figure}

\begin{figure}[htbp]
    \centering
    \includegraphics[width=\linewidth]{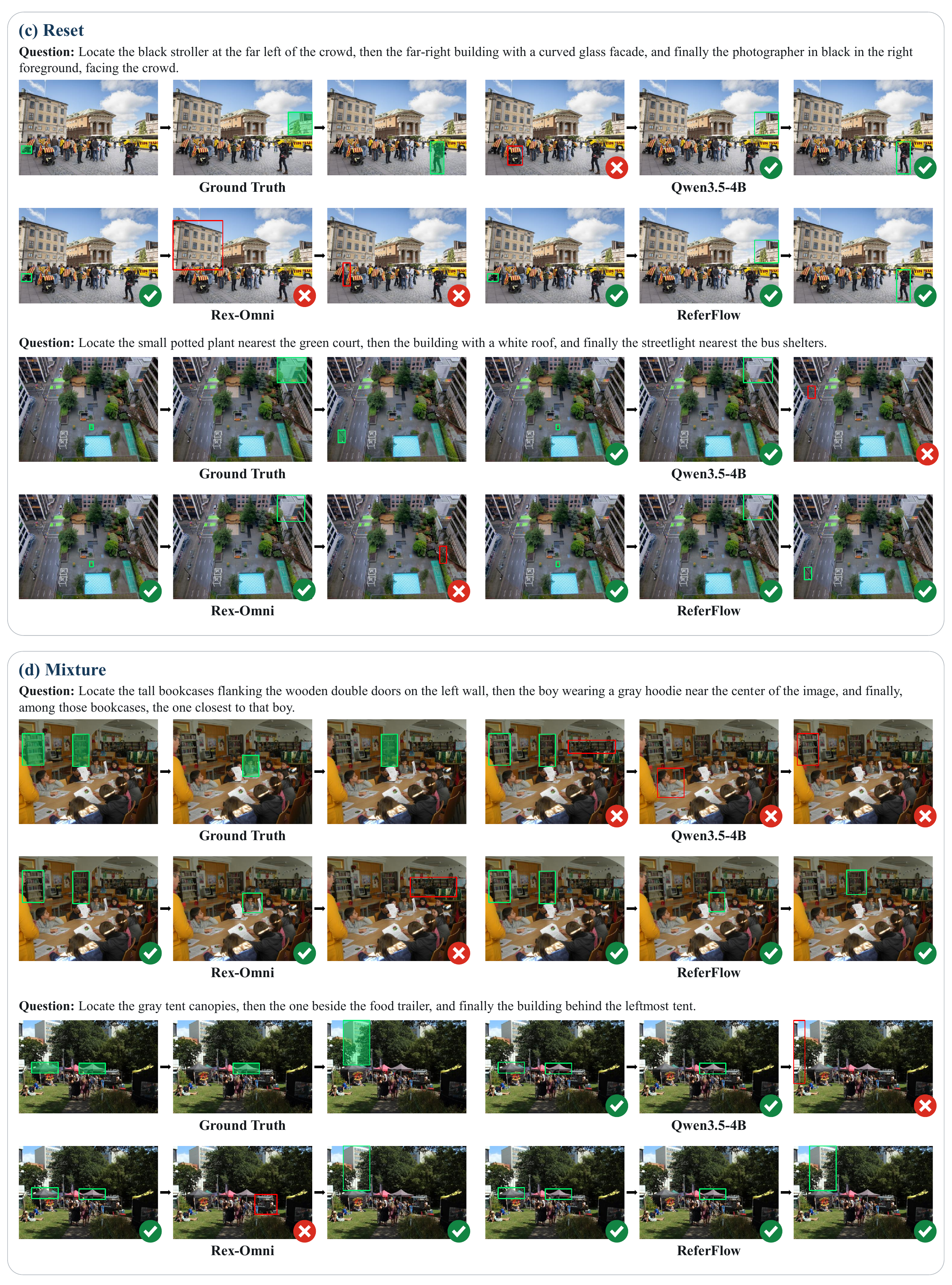}
    \caption{\textbf{Visualization results for Reset and Mixture on ReferFlow-Bench.}
Each panel presents a challenging example and the corresponding predictions from representative models.}
    \label{fig:bench_visualizations_resetmixture}
\end{figure}

We provide visualizations on ReferFlow-Bench in Fig.~\ref{fig:bench_visualizations_refineshift} and Fig.~\ref{fig:bench_visualizations_resetmixture}, covering all four trajectory types, and showing the predictions of representative models to illustrate the challenge of ReferFlow-Bench.


\subsection{Visualization Results of ReferFlow}
\begin{figure}[htbp]
    \centering
    \includegraphics[width=0.95\linewidth]{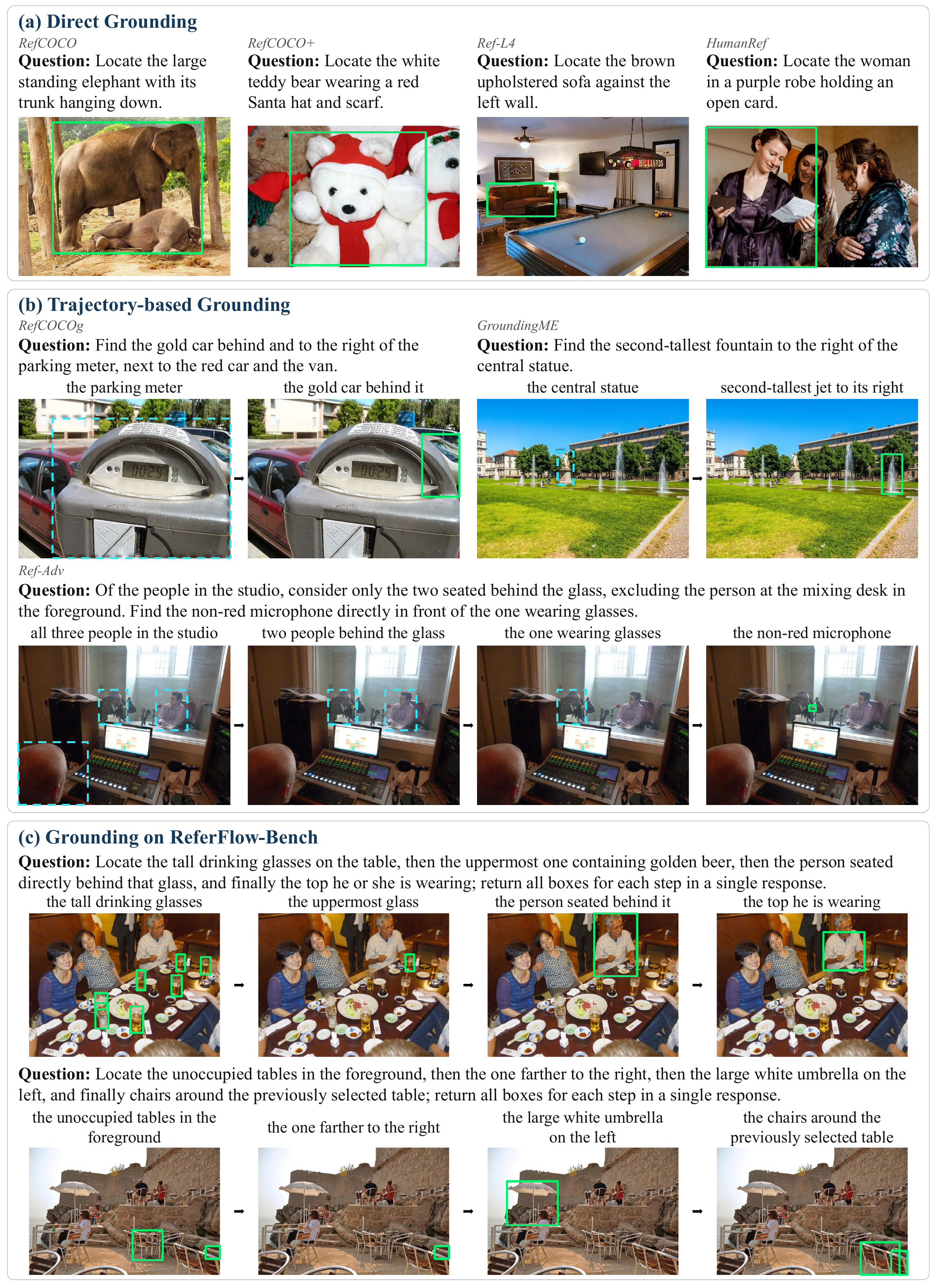}
    \caption{\textbf{Visualization results of ReferFlow.}
    Blue dashed boxes denote intermediate reference regions, while green solid boxes denote predicted target regions.
    }
    \label{fig:referflow_cases}
\end{figure}

Fig.~\ref{fig:referflow_cases} presents representative outputs of ReferFlow.
For REC tasks, ReferFlow adapts to query complexity by directly localizing targets or generating intermediate grounding steps.
On ReferFlow-Bench, it follows the provided grounding steps and consistently localizes the corresponding target regions.

\end{document}